\documentclass{article}

\usepackage[preprint]{neurips_2026}

\usepackage[utf8]{inputenc} 
\usepackage[T1]{fontenc}    
\usepackage{hyperref}       
\usepackage{url}            
\usepackage{booktabs}       
\usepackage{amsfonts}       
\usepackage{nicefrac}       
\usepackage{wrapfig}
\usepackage{microtype}      
\usepackage{xcolor}   
\usepackage{colortbl}
\usepackage{amsmath}
\usepackage{amssymb}
\usepackage{amsthm}
\newtheorem{theorem}{Theorem}
\usepackage{graphicx}
\usepackage{multirow}
\usepackage{makecell}
\usepackage{fontawesome5}
\newcolumntype{P}[1]{>{\centering\arraybackslash}p{#1}}
\definecolor{vlmgray}{RGB}{226, 226, 232}
\theoremstyle{remark}
\newtheorem{remark}{Remark}[section]
\usepackage{minitoc}
\mtcsetdepth{parttoc}{3}
\mtcsettitle{parttoc}{Appendix contents}

\title{TACT-KV: Tri-Axis Cosine Transform for Compressing Volumetric KV Caches in Medical VLMs}

\author{
  Chongyu Qu\textsuperscript{1} 
  \quad Ritchie Zhao\textsuperscript{3}
  \quad Yufan He\textsuperscript{3} 
  \quad Dong Yang\textsuperscript{3}
  \quad Zhengyi Lu\textsuperscript{1}
  \quad Junchao Zhu\textsuperscript{1}
  \\
  \bf
  Tianyuan Yao\textsuperscript{1}
  \quad Juming Xiong\textsuperscript{1}
  \quad Junlin Guo\textsuperscript{1}
  \quad Yanfan Zhu\textsuperscript{1}
  \quad Yuechen Yang\textsuperscript{1}
  \\
  \bf
  \quad Daguang Xu\textsuperscript{3}
  \quad Bennett A. Landman\textsuperscript{1,2}
  \quad Yucheng Tang\textsuperscript{3}
  \quad Yuankai Huo\textsuperscript{1,}\thanks{Corresponding author: Yuankai Huo (\href{mailto:yuankai.huo@vanderbilt.edu}{\texttt{yuankai.huo@vanderbilt.edu}}).}\\[0.5em]
  {\normalfont\textsuperscript{1}Vanderbilt University}\\
  {\normalfont\textsuperscript{2}Vanderbilt University Medical Center}\\
  {\normalfont\textsuperscript{3}NVIDIA}\\
  [0.5em]
  {\normalfont\small
  \href{https://github.com/Chongyu1117/TACT-KV}{%
    \faGithub\enspace
    \texttt{https://github.com/Chongyu1117/TACT-KV}}}
}

\begin{document}

\maketitle

\doparttoc 
\faketableofcontents 

\begin{abstract}
Vision-language models (VLMs) applied to 3D medical scans face a structural memory wall. A single CT or MRI volume produces tens of thousands of vision tokens, and the resulting key-value (KV) cache scales linearly with both batch and context. Existing KV compression methods operate on the cache's feature or token statistics, or apply 1D and 2D transforms, leaving the three-axis spatial structure in the cache unexploited. We propose \textbf{TACT-KV} (\underline{T}ri-\underline{A}xis \underline{C}osine \underline{T}ransform), a calibration-free KV compression method that exploits strong spatial correlations in visual KV states along all three anatomical axes of 3D scans. This correlation concentrates the KV cache energy under a frequency-domain transform. TACT-KV applies a separable 3D discrete cosine transform to the KV cache, allocates a per-layer bit budget across frequency bins via reverse water-filling (more bits to high-energy components), and quantizes each bin with a fixed scalar codebook. We derive a closed-form relation: the distortion reduction from adaptive bit allocation is dominated by the spectral concentration of the transformed cache.
Across six medical VLMs and two 3D CT benchmarks, TACT-KV achieves the lowest 1-bit prediction-flip rate on eleven of twelve model and dataset pairs in our main comparison. The same advantage reproduces on four general-purpose VLMs evaluated on video tasks, indicating that the benefit of exploiting three-axis structure generalizes beyond medical imaging. At equal high-bandwidth memory (HBM), TACT-KV scales the decoding batch size on a single GPU by up to $8\times$.
\end{abstract}

\section{Introduction}

Vision-language models (VLMs) are increasingly used for 3D medical image analysis~\cite{luo2023biomedgpt}, with applications spanning computed tomography (CT) report generation, magnetic resonance imaging (MRI) anomaly detection, and multimodal diagnostic assistance~\cite{sellergren2025medgemma, jiang2025hulu, li2023llava}. Unlike 2D natural images, a 3D medical scan contains hundreds of slices that capture fine-grained anatomical detail such as small lesions, subtle nodules, and vascular structures driving clinical diagnosis~\cite{hamamci2025better, blankemeier2024merlin, bai2024m3d}. The scan then produces tens of thousands of vision tokens after the vision encoder, and the resulting key-value (KV) cache scales linearly with both batch size and sequence length. Clinical sites typically operate on single-card or small-cluster GPUs~\cite{lan2026performance}, where this cache quickly dominates GPU memory and becomes the bottleneck during inference. Compressing the visual KV cache without sacrificing diagnostic fidelity is therefore a gating problem for clinical deployment.


KV cache compression directly addresses this bottleneck, and existing methods split into two categories by their target. (1) LLM-targeted methods, including KIVI~\cite{liu2024kivi}, KVQuant~\cite{hooper2024kvquant}, and TurboQuant~\cite{zandieh2025turboquant}, quantize the cache along its 1D token or channel axis. (2) VLM-targeted methods, including AKVQ-VL~\cite{su2025akvq} and VL-Cache~\cite{tu2024vl}, exploit attention-derived saliency to allocate precision or evict cache entries. A separate line of work instead transforms the cache into the frequency domain before compression. FreqKV~\cite{kai2025freqkv} applies a 1D discrete cosine transform (DCT) along the token sequence, FlashCache~\cite{yang2025revisiting} extracts a low-frequency baseline and handles outliers via IDCT and uses the reconstruction residual to identify outlier KV pairs for retention, KVTC~\cite{staniszewski2025kv} performs feature-dimension PCA with adaptive bit allocation, and Fourier-VLM~\cite{wang2025fourier} applies a 2D DCT to vision encoder features before they enter the cache. None of these methods exploit the spatial continuity that 3D scans carry along all three axes.

A closer analog is video KV cache compression, which exploits redundancy between adjacent 2D frames~\cite{sai2026dual, wan2024look, tao2025plug}. Medical VLMs commonly treat a 3D scan as a sequence of 2D slices~\cite{sellergren2025medgemma,jiang2025hulu}, but the slice axis differs from the temporal axis. The temporal axis is motion-induced, with attention concentrated on a few salient frames, while the slice axis is anatomically continuous, with diagnostic signal spread across many slices rather than localized in a sparse subset. Directly applying video-domain compression to medical scans is therefore questionable.
\begin{figure}[t]
\centering
\includegraphics[width=\linewidth]{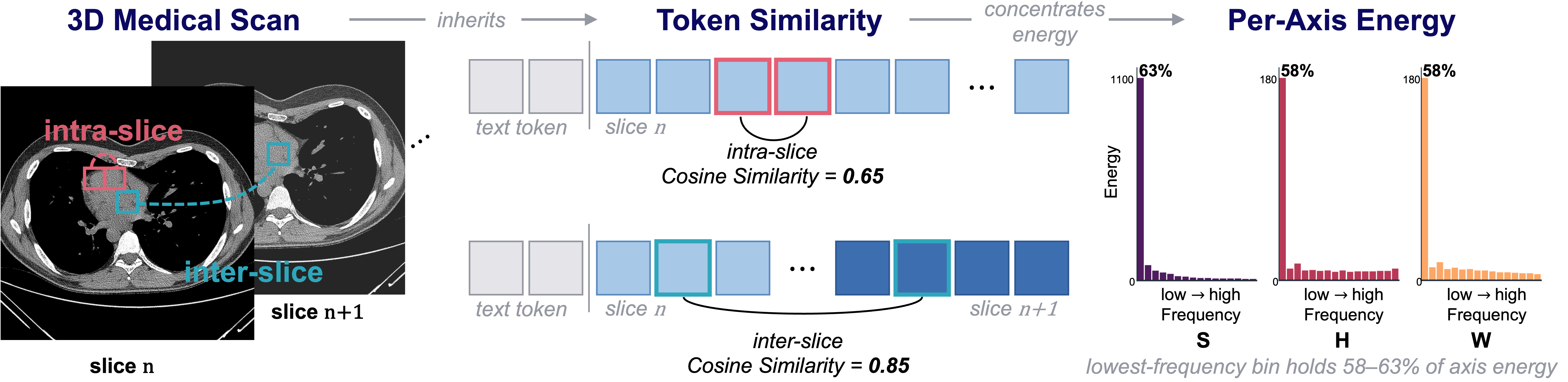}
\caption{\textbf{Motivation.} Tokens at the same in-plane spatial position across adjacent slices in the visual KV cache exhibit high pairwise similarity, and within each slice neighboring tokens are also strongly correlated. The cache energy then concentrates in a small set of low-frequency components along every axis, a three-axis structure that previous work leaves unexploited.}
\label{fig:motivation}
\end{figure}

3D medical scans carry a distinctive structural property. Nearby slices are highly similar, just like neighboring pixels within a single image. We observe that this structure also appears in the visual KV cache of the language model decoder: \textit{tokens at the same spatial location across adjacent slices have high similarity (Fig.~\ref{fig:motivation}), and the cache shows strong autocorrelation along all three spatial axes.} This spatial correlation concentrates the cache energy in a small set of low-frequency components, which a frequency-domain transform can compress.

Motivated by this observation, we propose \textbf{TACT-KV} (\underline{T}ri-\underline{A}xis \underline{C}osine \underline{T}ransform for KV Cache). TACT-KV reshapes the visual KV cache into a 3D tensor with dimensions (slice, height, width), then applies a separable 3D DCT across these three spatial axes. After transforming the cache into the frequency domain, TACT-KV allocates bits to different frequency components using a reverse water-filling strategy and compresses them with standard scalar quantization. The method requires no calibration data and introduces no additional learnable parameters.

We further prove that the gain of water-filling over uniform bit allocation grows as the cache energy concentrates in fewer frequency components, and verify this empirically. We evaluate TACT-KV on six medical VLMs using two 3D CT benchmarks, RSNA-ICH~\cite{flanders2020construction} and CT-RATE~\cite{hamamci2024developing}. Even at 1-bit quantization, TACT-KV performs close to the uncompressed cache, while existing LLM and VLM compression methods often fail. We also test TACT-KV on Video-MME~\cite{fu2025video, jiang2025hulu} using four general-purpose VLMs and observe similar improvements, suggesting that the method generalizes beyond medical imaging. At the same GPU memory budget, TACT-KV can increase the maximum decoding batch size by up to 8$\times$. Our contributions are:


\begin{itemize}
    \item We propose TACT-KV, to our knowledge the first method to treat the visual KV cache as a volumetric tensor and exploit the three-axis spatial correlation from 3D scans.
    \item We prove a closed-form expression for the distortion gain of water-filling over uniform bit allocation in terms of the spectral concentration of the cache, which we verify empirically.
    \item We evaluate TACT-KV across six medical VLMs (4B-32B), two 3D CT benchmarks, and four general-purpose VLMs on Video-MME. TACT-KV achieves the lowest 1-bit flip rate on eleven of twelve medical VLM and dataset pairs in our main comparison, while delivering up to $8\times$ larger decoding batch at equal HBM.
\end{itemize}

\begin{figure}[t]
  \centering
  \includegraphics[width=\textwidth]{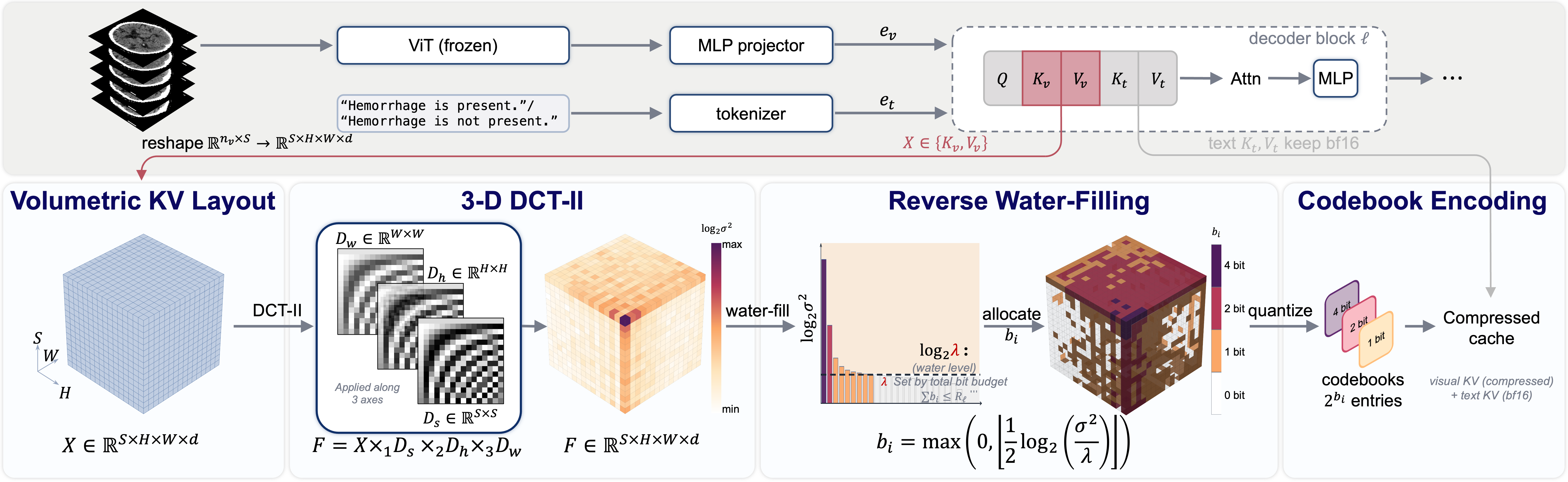}
  \caption{%
    \textbf{TACT-KV pipeline.}
  A medical VLM encodes a volumetric CT scan and a clinical prompt into visual and text embeddings $\mathbf{e}_v, \mathbf{e}_t$. Inside each decoder block, visual keys/values $K_v, V_v$ enter the four-stage pipeline below, while text $K_t, V_t$ bypass at bf16.
  \textbf{Volumetric KV Layout:} the visual cache is reshaped into a $(S, H, W)$ grid inherited from the vision encoder.
  \textbf{3-D DCT-II:} a separable transform along the three axes concentrates energy at the DC corner (magma colormap, $\log_2 \sigma^2$).
  \textbf{Reverse Water-Filling:} a per-layer bit budget is allocated above water level $\lambda$, with bins below dropped (bar color = bit width 4/2/1/0).
  \textbf{Codebook Encoding:} surviving bins are gain-shape encoded against the corresponding Lloyd--Max codebook (1/2/4-bit tiles shown).
  All spectral statistics are estimated on the fly, requiring no calibration data.%
  }
  \label{fig_method}
\end{figure}
%
%
%

\section{Method}
\label{sec:method}

We introduce TACT-KV, a calibration-free method for compressing the visual portion of the KV cache held by a decoder-only vision--language model during generation. For a backbone with $L$ transformer layers and $n_h$ attention heads per layer, a single sequence at layer $\ell$ produces a key tensor $K^{(\ell)}$ and a value tensor $V^{(\ell)}$ with per-head feature dimension $d$. In a vision--language model, tokens originate from two sources, a vision encoder and a text tokenizer. When the input is a volumetric medical scan or a video, visual tokens dominate the count and inherit a known three-dimensional grid $S \times H \times W$ from the vision encoder, where $S$ is the slice or temporal axis token count and $H, W$ are the spatial pooling dimensions. We exploit this structural prior by allocating the per-layer bit budget for the visual KV cache in a frequency-domain representation of this grid, retaining the information required by downstream tasks at low bit precision. The pipeline introduces no learned parameters and runs once per layer per sequence in a single forward pass. Figure~\ref{fig_method} illustrates the four stages.

\subsection{Volumetric KV Layout}
\label{sec:method:layout}

Let $K_v \in \mathbb{R}^{B \times n_h \times n_v \times d}$ denote the visual portion of the layer-$\ell$ key cache and $K_t \in \mathbb{R}^{B \times n_h \times n_t \times d}$ denote the text portion, with $B$ the batch dimension, $n_v = S \cdot H \cdot W$ the visual token count, and $n_t$ the text token count. The value tensors $V_v, V_t$ are defined analogously. The grid parameters $(S, H, W)$ are emitted by the vision encoder for each input and are read directly by our pipeline. We do not estimate any per-sample structural quantity. We reshape the visual key tensor into the volumetric grid:
\begin{equation}
\label{eq:reshape}
X = \mathrm{reshape}(K_v) \in \mathbb{R}^{B \times n_h \times S \times H \times W \times d},
\end{equation}
and reshape the visual value tensor identically. The remaining pipeline (3-D DCT, water-filling allocation, codebook encoding) runs independently on $K_v$ and $V_v$ with separate per-bin variance estimates and bit budgets. The batch and head dimensions $(B, n_h)$ act as outer indices for the remainder of the pipeline and are omitted from subsequent equations. The text portion $K_t, V_t$ is held at bf16 (bfloat16)~\cite{kalamkar2019study} throughout the pipeline.

\subsection{3-D DCT-II}
\label{sec:method:dct}

The discrete cosine transform of type II is an orthonormal energy-compacting transform~\cite{ahmed1974discrete}. For a real sequence of length $M$, its $M \times M$ matrix $D$ has entries indexed by output frequency $k$ and input sample $m$,
\begin{equation}
\label{eq:dct}
D[k, m] = \alpha_k \cos\!\left[\frac{\pi}{M}\!\left(m + \tfrac{1}{2}\right) k\right], \qquad 0 \le k, m \le M - 1,
\end{equation}
with normalization constants $\alpha_0 = \sqrt{1/M}$ and $\alpha_{k \ge 1} = \sqrt{2/M}$ chosen so that $D D^{\!\top} = I$. For a first-order Markov source, the DCT-II asymptotically approaches the Karhunen--Lo\`eve transform, and variants of it underpin the block transform in JPEG and successor image and video codecs~\cite{wallace1991jpeg}.

We extend the DCT-II to the reshaped tensor $X \in \mathbb{R}^{S \times H \times W \times d}$. Let $D_s \in \mathbb{R}^{S \times S}$, $D_h \in \mathbb{R}^{H \times H}$, and $D_w \in \mathbb{R}^{W \times W}$ be the DCT-II matrices on the corresponding axis lengths, each constructed by~\eqref{eq:dct}. The frequency-domain tensor $F \in \mathbb{R}^{S \times H \times W \times d}$ is defined by the mode product~\cite{kolda2009tensor}:
\begin{equation}
\label{eq:modeprod}
F = X \times_1 D_s \times_2 D_h \times_3 D_w,
\end{equation}
where $\times_n$ denotes matrix multiplication along the $n$-th tensor axis. The separable three-axis structure reduces the per-layer cost from $\mathcal{O}(n_v^2 \cdot d)$ for a non-separable transform to $\mathcal{O}\!\big((S + H + W) \cdot n_v \cdot d\big)$, and the parameter count from $(SHW)^2$ to $S^2 + H^2 + W^2$. The feature axis $d$ is left untransformed and is handled by the scalar quantizer (SQ) in Section~\ref{sec:method:codebook}. Because $D_s, D_h, D_w$ are each orthonormal, \eqref{eq:modeprod} is a lossless invertible change of variable that preserves the $\ell_2$ norm. Compression is introduced in the next subsection by allocating the per-layer bit budget across frequency bins.

\subsection{Reverse Water-Filling Bit Allocation}
\label{sec:method:waterfill}

Let the per-bin variance of the frequency tensor in~\eqref{eq:modeprod}, averaged over the batch, head, and feature dimensions, be
\begin{equation}
\label{eq:variance}
\sigma_i^2 = \mathbb{E}_{B,\, n_h,\, d}\!\left[F_i^2\right], \qquad i = 1, \ldots, N, \quad N = S H W.
\end{equation}
Given a per-layer bit budget $R_\ell$, our goal is to allocate non-negative integer bits $b_i$ to each frequency bin so as to minimize the total reconstruction MSE. Under the high-rate scalar-quantization distortion model for a Gaussian source~\cite{gersho2012vector}, the per-bin distortion is $\mathcal{D}_i(b_i) = c \cdot \sigma_i^2 \cdot 2^{-2 b_i}$ with $c$ a constant determined by the quantizer geometry~\cite{cover1991entropy}, and the problem reduces to
\begin{equation}
\label{eq:opt}
\min_{b_i \in \mathbb{Z}_{\ge 0}} \sum_i \sigma_i^2 \cdot 2^{-2 b_i}, \quad \text{s.t.} \quad \sum_i b_i \le R_\ell.
\end{equation}
A Lagrangian relaxation with a non-negativity constraint yields the reverse water-filling solution
\begin{equation}
\label{eq:waterfill}
b_i = \max\!\left(0,\; \left\lfloor \tfrac{1}{2} \log_2 (\sigma_i^2 / \lambda) \right\rfloor\right),
\end{equation}
where the water level $\lambda$ is determined by a bisection search on the budget equation. The $\max(0,\cdot)$ in~\eqref{eq:waterfill} enforces non-negativity, the floor produces integer bits, and a greedy $\pm 1$ correction enforces $\sum_i b_i = R_\ell$ exactly. In our implementation $b_i$ is capped at $8$.

The following theorem expresses the distortion improvement of reverse water-filling over uniform allocation as a closed-form function of the spectral shape. Let $\mathcal{D}_{\mathrm{uniform}}$ and $\mathcal{D}_{\mathrm{water\text{-}fill}}$ denote the total distortion summed over all bins under the two allocations.

\begin{theorem}
\label{thm:water_fill}
Let $\{\sigma_i^2\}_{i=1}^N$ be the per-bin variances in~\eqref{eq:variance}, and define the spectral flatness
\begin{equation}
\label{eq:flatness}
\gamma = \frac{\left(\prod_i \sigma_i^2\right)^{1/N}}{(1/N)\, \sum_i \sigma_i^2} \in (0, 1].
\end{equation}
Under matched bit budget $R_\ell$ and the high-rate Gaussian distortion model,
\begin{equation}
\label{eq:thm_ratio}
\mathcal{D}_{\mathrm{uniform}} \,/\, \mathcal{D}_{\mathrm{water\text{-}fill}} = \frac{(1/N) \sum_i \sigma_i^2}{\left(\prod_i \sigma_i^2\right)^{1/N}} = \frac{1}{\gamma}.
\end{equation}
\end{theorem}

\begin{proof}[Sketch]
Substituting $b_i = R_\ell/N$ into the distortion model gives $\mathcal{D}_{\mathrm{uniform}} = c \cdot 2^{-2 R_\ell / N} \cdot \sum_i \sigma_i^2$. Substituting the continuous solution of~\eqref{eq:waterfill} into the per-bin distortion gives $\mathcal{D}_i = c \lambda$, independent of $i$, so $\mathcal{D}_{\mathrm{water\text{-}fill}} = c N \lambda$. The budget equation $\sum_i b_i = R_\ell$ then yields $\lambda = \left(\prod_i \sigma_i^2\right)^{1/N} \cdot 2^{-2 R_\ell / N}$. Dividing the two expressions gives~\eqref{eq:thm_ratio}. The full proof appears in Appendix~\ref{app:proof}.
\end{proof}

\begin{wrapfigure}{r}{0.5\textwidth}
\centering
\includegraphics[width=0.5\textwidth]{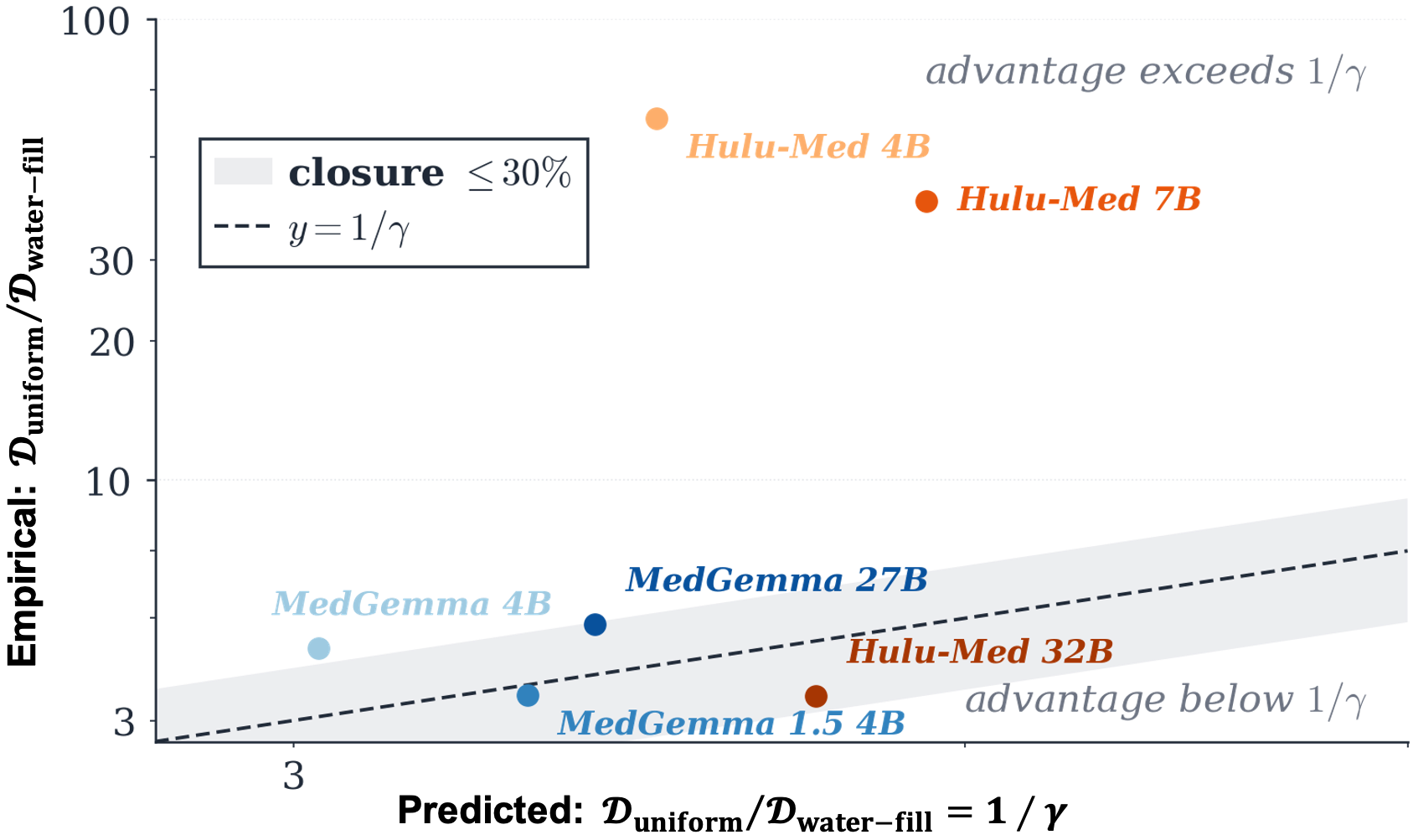}
\caption{\textbf{Theorem~\ref{thm:water_fill} closure on RSNA-ICH.} Empirical $\mathcal{D}_{\mathrm{uniform}} / \mathcal{D}_{\mathrm{water\text{-}fill}}$ versus closed-form $1/\gamma$ at $1$-bit average rate ($N = 3000$ studies, six VLMs). Shaded band marks $\leq 30\%$ deviation from $y = 1/\gamma$.}
\label{fig:closure}
\end{wrapfigure}

Theorem~\ref{thm:water_fill} expresses $1/\gamma$ as the closed-form distortion improvement of reverse water-filling over uniform allocation.\footnote{The result applies to the continuous relaxation that drops the floor in~\eqref{eq:waterfill}, assumes the high-rate Gaussian distortion model, and implicitly assumes that every bin receives positive rate so that $\mathcal{D}_{\mathrm{water\text{-}fill}} = c N \lambda$. The integer floor and the $\pm 1$ correction contribute a finite-rate penalty to the water-filling distortion, and both the high-rate model and the all-active assumption lose validity as the average bit rate per coordinate approaches zero.} As $\gamma \to 1$, the spectrum is flat and water-filling reduces to uniform allocation. As $\gamma \to 0$, energy concentrates in a small subset of bins and the closed-form $1/\gamma$ grows without bound. In practice the integer water-filling in~\eqref{eq:waterfill} drops the inactive bins and the realized gain is finite. Figure~\ref{fig:closure} compares this closed-form prediction to the empirical noise-variance ratio on RSNA-ICH. Three of six cells fall within the $30\%$ closure band, one cell sits at $40\%$ (directional agreement), and two cells (Hulu-Med 4B and 7B) lie above the identity line in the low-rate regime where the high-rate Gaussian distortion model underlying Theorem~\ref{thm:water_fill} relaxes. When noise or motion weakens spatial correlation and flattens the cache spectrum, $\gamma$ approaches one and the predicted allocation gain decreases.

\subsection{Codebook Encoding}
\label{sec:method:codebook}

The integer bit allocation $b_i$ from~\eqref{eq:waterfill} routes each frequency bin to a scalar quantizer of matching width. We use a gain-shape scalar quantizer as the backend. For each bin with $b_i \ge 1$, we extract the $\ell_2$ norm $g$, normalize the residual onto the unit sphere, apply a fixed orthogonal rotation shared across all bins~\cite{chee2023quip}, and look up each rotated coordinate against a precomputed Lloyd--Max boundary table~\cite{lloyd1982least} at width $b_i$. The total per-layer cache occupancy is $\sum_{i:\, b_i \ge 1} (b_i \cdot d + 32)$ bits plus a small index overhead, where the fp32 scalar $g$ contributes the constant $32$ for each non-empty bin. We compare gain-shape SQ against symmetric uniform and NormalFloat backends in Section~\ref{sec:experiments}. Decoding inverts the steps above. We read the indices, dequantize, undo the rotation, multiply by $g$, and apply the transpose mode product to recover $\hat{X} = \hat{F} \times_1 D_s^{\!\top} \times_2 D_h^{\!\top} \times_3 D_w^{\!\top}$. Both the inverse DCT and the codebook dequantization are deterministic and remain transparent to the attention implementation. A fused implementation of the inverse path is described in Appendix~\ref{app:kernel}.

\section{Experiments and Results}
\label{sec:experiments}

\subsection{Experimental Setup}
\label{sec:exp:setup}

\smallskip\noindent\textbf{Datasets and VLMs.}
We evaluate on three benchmarks across ten VLMs in two domains. For \textbf{medical CT}, we use RSNA-ICH~\cite{flanders2020construction} on a stratified $N{=}3{,}000$ subset of the 2019 training split under the standard six-label hemorrhage protocol, and CT-RATE~\cite{hamamci2024developing} on the full validation split ($N{=}3{,}039$ volumes) under the official $18$-abnormality multi-label protocol. We evaluate six medical VLMs on these benchmarks: MedGemma 1.5/4B/27B~\cite{sellergren2025medgemma} and Hulu-Med 4B/7B/32B~\cite{jiang2025hulu}. For \textbf{cross-domain transfer}, we use Video-MME~\cite{fu2025video} in the no-subtitle setting on all $2{,}700$ four-way multiple-choice questions (MCQs), with four general-purpose VLMs: Qwen2.5-VL-7B~\cite{bai2025qwen25vltechnicalreport}, Qwen3-VL-8B~\cite{bai2025qwen3}, LLaVA-OneVision-7B~\cite{li2024llava}, and VideoLLaMA3-7B~\cite{zhang2025videollama}. The lineup spans three VL variants, three vision encoders, and four LLM versions (Qwen2, Qwen2.5, Qwen3, Gemma-3).

\smallskip\noindent\textbf{Implementation.}
All experiments run on NVIDIA RTX PRO 6000 (96 GiB) GPUs. For the medical benchmarks we use rank classification on assertion-style completions~\cite{gao2021framework}; for Video-MME we use the four-letter MCQ prompt with greedy decoding. All methods compress only the visual portion of the cache and use the same inputs and prompts, with non-visual tokens (chat template, boundary markers) held at bf16. This isolates the contribution of our method and is consistent with prior VLM-targeted compression that treats visual and text tokens separately~\cite{su2025akvq,tao2025plug,tu2024vl}. We report bit budgets $\{1, 2, 3, 4\}$, with primary focus on the 1-bit setting where compression methods diverge most sharply. The $8$-bit setting is approximately lossless across all backends~\cite{liu2024kivi,hooper2024kvquant} and is omitted. Full per-VLM configurations and protocols are in Appendix~\ref{app:impl}.

\smallskip\noindent\textbf{Metrics.}
The primary metric is the \textbf{macro prediction-flip rate} relative to the bf16 reference, the fraction of per-study per-label binary predictions whose label changes under quantization, averaged over labels. Flip rate measures consistency with the bf16 decision boundary, a desirable property when a deployed system swaps in a compressed cache. We complement flip rate with macro AUROC and $95\%$ bootstrap CIs. For Video-MME we report per-question accuracy on the four-way MCQ.

\subsection{Main Results: Medical CT}
\label{sec:exp:main}

\begin{table*}[t]
\caption{\textbf{Macro prediction-flip rate (\%) relative to bf$16$ on RSNA-ICH and CT-RATE.} Entries include Wald $95\%$ CI half-widths. \textbf{Bold} and \underline{underline} mark the lowest and second-lowest measured values in each column. RSNA-ICH uses a stratified subset of $N{=}3{,}000$ studies under the six-label challenge protocol~\cite{flanders2020construction}. Each CT-RATE entry aggregates $54{,}702$ predictions from $N{=}3{,}039$ official validation volumes under the $18$-abnormality protocol~\cite{hamamci2024developing}. Bit budgets denote nominal quantizer rates and exclude metadata. For KVTC, the budget is the average number of bits per PCA coefficient before entropy coding.}
\label{tab:flip_medical}
\centering
\scriptsize
\setlength{\tabcolsep}{3pt}
\renewcommand{\arraystretch}{0.92}
\begin{tabular}{l *{4}{P{0.09\linewidth}} | *{4}{P{0.09\linewidth}}}
\toprule
& \multicolumn{4}{c|}{\textbf{RSNA-ICH} ($N{=}3{,}000$)} & \multicolumn{4}{c}{\textbf{CT-RATE} ($N{=}3{,}039$)} \\
\cmidrule(lr){2-5}\cmidrule(lr){6-9}
Method & 1-bit & 2-bit & 3-bit & 4-bit & 1-bit & 2-bit & 3-bit & 4-bit \\
\midrule

\rowcolor{vlmgray}\multicolumn{9}{c}{\textbf{\textit{MedGemma 1.5 4B}}}\\
TACT-KV (ours) & \textbf{3.78}\textsubscript{$\pm$0.28} & \textbf{1.91}\textsubscript{$\pm$0.20} & 1.29\textsubscript{$\pm$0.17} & 1.09\textsubscript{$\pm$0.15} & \textbf{4.69}\textsubscript{$\pm$0.18} & \textbf{2.33}\textsubscript{$\pm$0.13} & 1.89\textsubscript{$\pm$0.11} & 1.54\textsubscript{$\pm$0.10} \\
TurboQuant & \underline{6.27}\textsubscript{$\pm$0.35} & 4.47\textsubscript{$\pm$0.30} & 2.16\textsubscript{$\pm$0.21} & 1.41\textsubscript{$\pm$0.17} & 9.14\textsubscript{$\pm$0.24} & 4.61\textsubscript{$\pm$0.18} & 2.46\textsubscript{$\pm$0.13} & 1.87\textsubscript{$\pm$0.11} \\
KIVI & 10.74\textsubscript{$\pm$0.45} & \underline{1.96}\textsubscript{$\pm$0.20} & \textbf{1.12}\textsubscript{$\pm$0.15} & \underline{1.06}\textsubscript{$\pm$0.15} & 23.22\textsubscript{$\pm$0.35} & 4.66\textsubscript{$\pm$0.18} & \textbf{1.59}\textsubscript{$\pm$0.10} & \textbf{1.51}\textsubscript{$\pm$0.10} \\
KVQuant & 9.19\textsubscript{$\pm$0.42} & 2.31\textsubscript{$\pm$0.22} & \underline{1.26}\textsubscript{$\pm$0.16} & \textbf{1.02}\textsubscript{$\pm$0.15} & 9.46\textsubscript{$\pm$0.25} & \underline{3.49}\textsubscript{$\pm$0.15} & \underline{1.66}\textsubscript{$\pm$0.11} & \underline{1.51}\textsubscript{$\pm$0.10} \\
KVTC & 16.55\textsubscript{$\pm$0.54} & 6.93\textsubscript{$\pm$0.37} & 1.53\textsubscript{$\pm$0.18} & 2.26\textsubscript{$\pm$0.22} & 37.75\textsubscript{$\pm$0.41} & 17.91\textsubscript{$\pm$0.32} & 7.04\textsubscript{$\pm$0.21} & 2.86\textsubscript{$\pm$0.14} \\
VidKV & 7.54\textsubscript{$\pm$0.39} & 2.92\textsubscript{$\pm$0.25} & -- & -- & \underline{9.13}\textsubscript{$\pm$0.24} & 6.01\textsubscript{$\pm$0.20} & -- & -- \\
AKVQ-VL & 28.54\textsubscript{$\pm$0.66} & 2.10\textsubscript{$\pm$0.21} & -- & -- & 20.72\textsubscript{$\pm$0.34} & 4.10\textsubscript{$\pm$0.17} & -- & -- \\

\rowcolor{vlmgray}\multicolumn{9}{c}{\textbf{\textit{MedGemma 4B}}}\\
TACT-KV (ours) & \underline{2.37}\textsubscript{$\pm$0.22} & 0.91\textsubscript{$\pm$0.14} & \textbf{0.64}\textsubscript{$\pm$0.12} & 0.67\textsubscript{$\pm$0.12} & \textbf{2.95}\textsubscript{$\pm$0.14} & \textbf{1.98}\textsubscript{$\pm$0.12} & \textbf{1.69}\textsubscript{$\pm$0.11} & \underline{1.57}\textsubscript{$\pm$0.10} \\
TurboQuant & 4.41\textsubscript{$\pm$0.30} & 1.21\textsubscript{$\pm$0.16} & 0.81\textsubscript{$\pm$0.13} & 0.64\textsubscript{$\pm$0.12} & \underline{7.33}\textsubscript{$\pm$0.22} & 3.97\textsubscript{$\pm$0.16} & 2.22\textsubscript{$\pm$0.12} & 2.20\textsubscript{$\pm$0.12} \\
KIVI & 2.43\textsubscript{$\pm$0.23} & \underline{0.84}\textsubscript{$\pm$0.13} & \underline{0.67}\textsubscript{$\pm$0.12} & \underline{0.63}\textsubscript{$\pm$0.12} & 11.58\textsubscript{$\pm$0.27} & 3.80\textsubscript{$\pm$0.16} & \underline{1.73}\textsubscript{$\pm$0.11} & \textbf{1.56}\textsubscript{$\pm$0.10} \\
KVQuant & 10.58\textsubscript{$\pm$0.45} & 1.22\textsubscript{$\pm$0.16} & 0.73\textsubscript{$\pm$0.12} & \textbf{0.61}\textsubscript{$\pm$0.11} & 7.92\textsubscript{$\pm$0.23} & \underline{2.84}\textsubscript{$\pm$0.14} & 2.09\textsubscript{$\pm$0.12} & 1.65\textsubscript{$\pm$0.11} \\
KVTC & 7.12\textsubscript{$\pm$0.38} & 24.77\textsubscript{$\pm$0.63} & 3.45\textsubscript{$\pm$0.27} & 1.38\textsubscript{$\pm$0.17} & 16.86\textsubscript{$\pm$0.31} & 13.67\textsubscript{$\pm$0.29} & 5.57\textsubscript{$\pm$0.19} & 2.17\textsubscript{$\pm$0.12} \\
VidKV & 14.13\textsubscript{$\pm$0.51} & 0.90\textsubscript{$\pm$0.14} & -- & -- & 15.09\textsubscript{$\pm$0.30} & 2.99\textsubscript{$\pm$0.14} & -- & -- \\
AKVQ-VL & \textbf{2.20}\textsubscript{$\pm$0.21} & \textbf{0.76}\textsubscript{$\pm$0.13} & -- & -- & 23.28\textsubscript{$\pm$0.35} & 3.58\textsubscript{$\pm$0.16} & -- & -- \\

\rowcolor{vlmgray}\multicolumn{9}{c}{\textbf{\textit{MedGemma 27B}}}\\
TACT-KV (ours) & \textbf{0.37}\textsubscript{$\pm$0.09} & \textbf{0.26}\textsubscript{$\pm$0.08} & \underline{0.22}\textsubscript{$\pm$0.07} & 0.28\textsubscript{$\pm$0.08} & \textbf{2.63}\textsubscript{$\pm$0.13} & \textbf{1.33}\textsubscript{$\pm$0.10} & \textbf{1.44}\textsubscript{$\pm$0.10} & \underline{1.38}\textsubscript{$\pm$0.10} \\
TurboQuant & 4.31\textsubscript{$\pm$0.30} & 0.47\textsubscript{$\pm$0.10} & 0.31\textsubscript{$\pm$0.08} & 0.24\textsubscript{$\pm$0.07} & \underline{9.14}\textsubscript{$\pm$0.24} & 3.76\textsubscript{$\pm$0.16} & 1.63\textsubscript{$\pm$0.11} & 1.64\textsubscript{$\pm$0.11} \\
KIVI & 16.37\textsubscript{$\pm$0.54} & 0.42\textsubscript{$\pm$0.09} & \textbf{0.21}\textsubscript{$\pm$0.07} & \underline{0.21}\textsubscript{$\pm$0.07} & 19.83\textsubscript{$\pm$0.33} & 2.26\textsubscript{$\pm$0.12} & \underline{1.61}\textsubscript{$\pm$0.11} & 1.38\textsubscript{$\pm$0.10} \\
KVQuant & 20.75\textsubscript{$\pm$0.59} & 0.38\textsubscript{$\pm$0.09} & 0.32\textsubscript{$\pm$0.08} & 0.25\textsubscript{$\pm$0.07} & 14.13\textsubscript{$\pm$0.29} & 2.73\textsubscript{$\pm$0.14} & 1.72\textsubscript{$\pm$0.11} & 1.49\textsubscript{$\pm$0.10} \\
KVTC & 50.57\textsubscript{$\pm$0.73} & 10.56\textsubscript{$\pm$0.45} & 0.27\textsubscript{$\pm$0.08} & \textbf{0.18}\textsubscript{$\pm$0.06} & 28.43\textsubscript{$\pm$0.38} & 6.88\textsubscript{$\pm$0.21} & 2.19\textsubscript{$\pm$0.12} & \textbf{0.96}\textsubscript{$\pm$0.08} \\
VidKV & \underline{2.79}\textsubscript{$\pm$0.24} & \underline{0.38}\textsubscript{$\pm$0.09} & -- & -- & 14.49\textsubscript{$\pm$0.29} & 2.10\textsubscript{$\pm$0.12} & -- & -- \\
AKVQ-VL & 15.48\textsubscript{$\pm$0.53} & 0.43\textsubscript{$\pm$0.10} & -- & -- & 12.48\textsubscript{$\pm$0.28} & \underline{1.72}\textsubscript{$\pm$0.11} & -- & -- \\

\rowcolor{vlmgray}\multicolumn{9}{c}{\textbf{\textit{Hulu-Med 7B}}}\\
TACT-KV (ours) & \textbf{6.08}\textsubscript{$\pm$0.35} & \underline{3.51}\textsubscript{$\pm$0.27} & 2.33\textsubscript{$\pm$0.22} & 1.86\textsubscript{$\pm$0.20} & \textbf{10.44}\textsubscript{$\pm$0.26} & \textbf{4.71}\textsubscript{$\pm$0.18} & \textbf{3.43}\textsubscript{$\pm$0.15} & \textbf{2.78}\textsubscript{$\pm$0.14} \\
TurboQuant & 36.25\textsubscript{$\pm$0.70} & 28.07\textsubscript{$\pm$0.66} & 40.44\textsubscript{$\pm$0.72} & 52.33\textsubscript{$\pm$0.73} & 52.95\textsubscript{$\pm$0.42} & 35.24\textsubscript{$\pm$0.40} & 43.01\textsubscript{$\pm$0.41} & 36.42\textsubscript{$\pm$0.40} \\
KIVI & 25.38\textsubscript{$\pm$0.64} & 4.37\textsubscript{$\pm$0.30} & \textbf{2.25}\textsubscript{$\pm$0.22} & \textbf{1.67}\textsubscript{$\pm$0.19} & 56.62\textsubscript{$\pm$0.42} & 8.53\textsubscript{$\pm$0.23} & \underline{4.77}\textsubscript{$\pm$0.18} & 3.48\textsubscript{$\pm$0.15} \\
KVQuant & \underline{8.29}\textsubscript{$\pm$0.40} & 4.10\textsubscript{$\pm$0.29} & \underline{2.32}\textsubscript{$\pm$0.22} & 1.90\textsubscript{$\pm$0.20} & \underline{16.82}\textsubscript{$\pm$0.31} & 7.38\textsubscript{$\pm$0.22} & 5.06\textsubscript{$\pm$0.18} & \underline{3.28}\textsubscript{$\pm$0.15} \\
KVTC & 60.38\textsubscript{$\pm$0.71} & 13.57\textsubscript{$\pm$0.50} & 3.12\textsubscript{$\pm$0.25} & \underline{1.75}\textsubscript{$\pm$0.19} & 48.76\textsubscript{$\pm$0.42} & 43.15\textsubscript{$\pm$0.42} & 6.38\textsubscript{$\pm$0.20} & 3.31\textsubscript{$\pm$0.15} \\
VidKV & 22.18\textsubscript{$\pm$0.61} & 4.02\textsubscript{$\pm$0.29} & -- & -- & 45.51\textsubscript{$\pm$0.42} & 8.19\textsubscript{$\pm$0.23} & -- & -- \\
AKVQ-VL & 32.76\textsubscript{$\pm$0.69} & \textbf{3.26}\textsubscript{$\pm$0.26} & -- & -- & 56.24\textsubscript{$\pm$0.42} & \underline{6.33}\textsubscript{$\pm$0.20} & -- & -- \\

\rowcolor{vlmgray}\multicolumn{9}{c}{\textbf{\textit{Hulu-Med 4B}}}\\
TACT-KV (ours) & \textbf{8.53}\textsubscript{$\pm$0.41} & \underline{3.91}\textsubscript{$\pm$0.28} & \underline{2.43}\textsubscript{$\pm$0.23} & \underline{2.05}\textsubscript{$\pm$0.21} & \textbf{7.22}\textsubscript{$\pm$0.22} & \textbf{3.70}\textsubscript{$\pm$0.16} & \underline{2.78}\textsubscript{$\pm$0.14} & \textbf{2.33}\textsubscript{$\pm$0.13} \\
TurboQuant & 76.86\textsubscript{$\pm$0.62} & 78.61\textsubscript{$\pm$0.60} & 19.47\textsubscript{$\pm$0.58} & 4.30\textsubscript{$\pm$0.30} & 39.06\textsubscript{$\pm$0.41} & 37.06\textsubscript{$\pm$0.40} & 27.99\textsubscript{$\pm$0.38} & 5.69\textsubscript{$\pm$0.19} \\
KIVI & 77.01\textsubscript{$\pm$0.61} & 4.65\textsubscript{$\pm$0.31} & \textbf{2.37}\textsubscript{$\pm$0.22} & \textbf{2.00}\textsubscript{$\pm$0.20} & 37.03\textsubscript{$\pm$0.40} & 4.77\textsubscript{$\pm$0.18} & \textbf{2.66}\textsubscript{$\pm$0.13} & \underline{2.37}\textsubscript{$\pm$0.13} \\
KVQuant & \underline{28.85}\textsubscript{$\pm$0.66} & 6.51\textsubscript{$\pm$0.36} & 2.80\textsubscript{$\pm$0.24} & 2.16\textsubscript{$\pm$0.21} & \underline{17.17}\textsubscript{$\pm$0.32} & 5.70\textsubscript{$\pm$0.19} & 3.02\textsubscript{$\pm$0.14} & 2.41\textsubscript{$\pm$0.13} \\
KVTC & 40.44\textsubscript{$\pm$0.72} & 75.53\textsubscript{$\pm$0.63} & 5.82\textsubscript{$\pm$0.34} & 3.28\textsubscript{$\pm$0.26} & 56.24\textsubscript{$\pm$0.42} & 27.44\textsubscript{$\pm$0.37} & 6.70\textsubscript{$\pm$0.21} & 2.94\textsubscript{$\pm$0.14} \\
VidKV & 30.52\textsubscript{$\pm$0.67} & 4.34\textsubscript{$\pm$0.30} & -- & -- & 24.68\textsubscript{$\pm$0.36} & 4.77\textsubscript{$\pm$0.18} & -- & -- \\
AKVQ-VL & 75.68\textsubscript{$\pm$0.63} & \textbf{3.70}\textsubscript{$\pm$0.28} & -- & -- & 36.13\textsubscript{$\pm$0.40} & \underline{3.71}\textsubscript{$\pm$0.16} & -- & -- \\

\rowcolor{vlmgray}\multicolumn{9}{c}{\textbf{\textit{Hulu-Med 32B}}}\\
TACT-KV (ours) & \textbf{9.13}\textsubscript{$\pm$0.42} & \underline{4.40}\textsubscript{$\pm$0.30} & \underline{3.05}\textsubscript{$\pm$0.25} & 2.71\textsubscript{$\pm$0.24} & \textbf{10.04}\textsubscript{$\pm$0.25} & \textbf{4.33}\textsubscript{$\pm$0.17} & \underline{2.75}\textsubscript{$\pm$0.14} & \underline{2.21}\textsubscript{$\pm$0.12} \\
TurboQuant & 24.91\textsubscript{$\pm$0.63} & 6.66\textsubscript{$\pm$0.36} & 4.02\textsubscript{$\pm$0.29} & 3.25\textsubscript{$\pm$0.26} & 22.24\textsubscript{$\pm$0.35} & 6.75\textsubscript{$\pm$0.21} & 3.64\textsubscript{$\pm$0.16} & 2.84\textsubscript{$\pm$0.14} \\
KIVI & 43.32\textsubscript{$\pm$0.72} & 5.89\textsubscript{$\pm$0.34} & \textbf{2.95}\textsubscript{$\pm$0.25} & \underline{2.51}\textsubscript{$\pm$0.23} & 65.09\textsubscript{$\pm$0.40} & 5.14\textsubscript{$\pm$0.18} & \textbf{2.71}\textsubscript{$\pm$0.14} & \textbf{2.06}\textsubscript{$\pm$0.12} \\
KVQuant & \underline{11.30}\textsubscript{$\pm$0.46} & 5.32\textsubscript{$\pm$0.33} & 3.22\textsubscript{$\pm$0.26} & 2.57\textsubscript{$\pm$0.23} & \underline{14.39}\textsubscript{$\pm$0.29} & 6.05\textsubscript{$\pm$0.20} & 2.98\textsubscript{$\pm$0.14} & 2.30\textsubscript{$\pm$0.13} \\
KVTC & 48.36\textsubscript{$\pm$0.73} & 23.02\textsubscript{$\pm$0.62} & 3.31\textsubscript{$\pm$0.26} & \textbf{2.46}\textsubscript{$\pm$0.23} & 52.08\textsubscript{$\pm$0.42} & 28.90\textsubscript{$\pm$0.38} & 3.51\textsubscript{$\pm$0.15} & 2.30\textsubscript{$\pm$0.13} \\
VidKV & 34.67\textsubscript{$\pm$0.70} & 5.56\textsubscript{$\pm$0.33} & -- & -- & 30.80\textsubscript{$\pm$0.39} & \underline{4.79}\textsubscript{$\pm$0.18} & -- & -- \\
AKVQ-VL & 35.31\textsubscript{$\pm$0.70} & \textbf{4.20}\textsubscript{$\pm$0.29} & -- & -- & 50.47\textsubscript{$\pm$0.42} & 5.46\textsubscript{$\pm$0.19} & -- & -- \\

\bottomrule
\end{tabular}
\end{table*}

Table~\ref{tab:flip_medical} reports macro prediction-flip rate on RSNA-ICH and CT-RATE for all six medical VLMs against six representative baselines.

\begin{figure}[h]
  \centering
  \includegraphics[width=\textwidth]{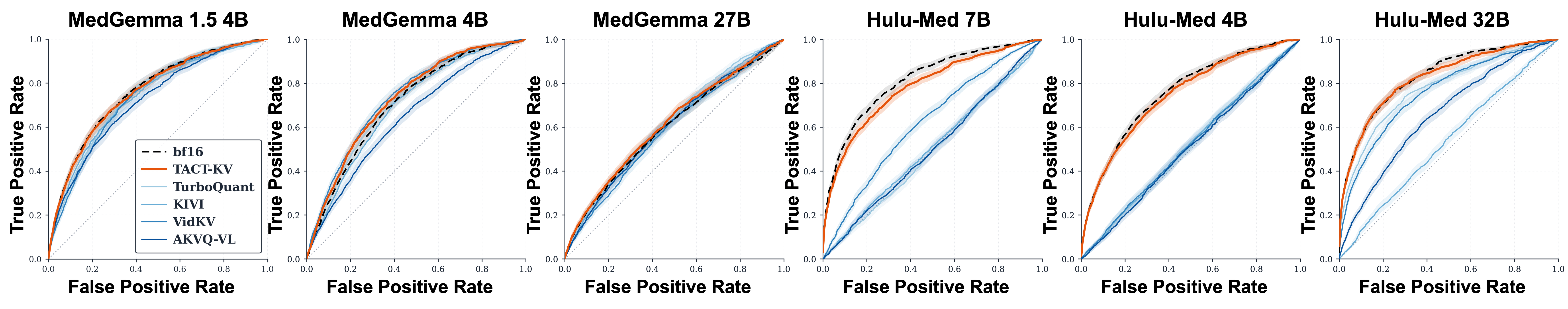}
  \caption{\textbf{Macro-averaged ROC curves at 1-bit on RSNA-ICH ($N{=}3{,}000$).} One panel per medical VLM. The black dashed line marks the bf16 reference; translucent bands are $95\%$ CIs from $1{,}000$ bootstrap resamples.}
  \label{fig:rsna_roc}
\end{figure}

\smallskip\noindent\textbf{TACT-KV achieves the lowest 1-bit flip rate on eleven of twelve VLM--dataset pairs.}
Across the eleven leading pairs, the next-best method has an average flip rate $2.60\times$ that of TACT-KV. Several baselines incur large prediction changes at 1-bit. KIVI's per-channel grouped-uniform primitive collapses on Hulu-Med 4B RSNA-ICH (77\% flip) and Hulu-Med 32B CT-RATE (65\%). TurboQuant exceeds 35\% flip on two of six VLMs. VidKV and AKVQ-VL reach worst-case 1-bit flip rates of 45.5\% and 75.7\% respectively. We attribute these failures to fixed quantization grids that, at 1-bit, lack the dynamic range to absorb the heavy-tailed visual KV distribution~\cite{dettmers2022gpt3,xiao2023smoothquant,lin2024awq,frantar2022gptq}. TACT-KV avoids this regime by allocating bits adaptively in the spectral domain, where the cache energy is concentrated.

\smallskip\noindent\textbf{TACT-KV stays within the bf16 confidence band on every VLM at 1-bit.}
Figure~\ref{fig:rsna_roc} shows macro-averaged ROC curves at 1-bit on RSNA-ICH. TACT-KV tracks the bf16 reference within its $95\%$ bootstrap CI on all six VLMs, while baseline curves drift outside the band on the Hulu-Med variants. The contrast is sharper on Hulu-Med than MedGemma because the Hulu-Med backbone uses full attention at every layer~\cite{jiang2025hulu}, exposing all layers to the $1$-bit codebook, whereas the Gemma-3 backbone interleaves five sliding-window layers per global layer~\cite{team2025gemma} and we leave the sliding-window layers at bf16 following the production convention of vLLM~\cite{kwon2023efficient}.

\subsection{Theorem~1 Closure}
\label{sec:exp:closure}

We extend the closure check of Theorem~\ref{thm:water_fill} from the six RSNA-ICH points in Figure~\ref{fig:closure} to all twelve\begin{wrapfigure}{r}{0.5\textwidth}
\centering
\includegraphics[width=0.49\textwidth]{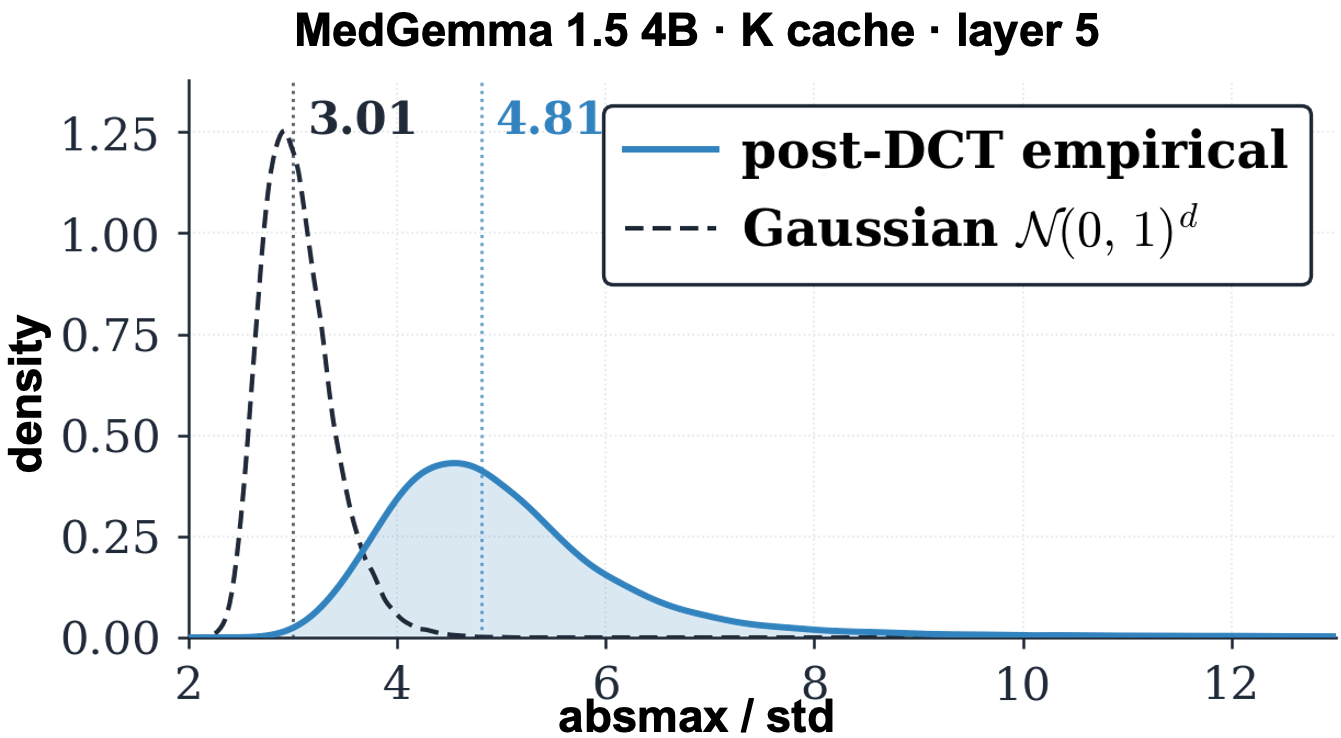}
\caption{\textbf{Post-DCT K-cache coefficient distribution is heavy-tailed.}}
\label{fig:quant_mechanism}
\end{wrapfigure} VLM--dataset pairs at 1-bit average rate. We isolate the allocation step by holding the 3D DCT and Gain-shape SQ (\S\ref{sec:method:codebook}) fixed and swapping reverse water-filling for uniform allocation. The empirical ratio $\hat{r}$ is computed from the per-study per-label logit-margin variance under the two allocations and serves as a first-order proxy for the cache distortion ratio $\mathcal{D}_{\text{uniform}}/\mathcal{D}_{\text{water-fill}}$ predicted by Theorem~\ref{thm:water_fill}. Table~\ref{tab:closure} shows that \textbf{eight of twelve pairs fall within $50\%$ of the closed-form $1/\gamma_K$}. The remaining four are Hulu-Med 4B and 7B on both datasets, where the 1-bit setting falls outside the high-rate regime. Results at 2 and 4 bits are provided in Appendix~\ref{app:cr_closure}.

\begin{table*}[t]
\caption{\textbf{Theorem~\ref{thm:water_fill} closure on twelve VLM-dataset pairs at 1-bit average rate.} For each pair we report the K-cache spectral flatness $\gamma_K$, the closed-form prediction $1/\gamma_K$, the empirical ratio $\hat{r}$, and the closure $|\hat{r} - 1/\gamma_K|/(1/\gamma_K)$. \textbf{Bold} marks closure within $30\%$ of $1/\gamma_K$, \underline{underline} within $50\%$. The four unmarked entries (Hulu-Med 4B and 7B) lie outside the high-rate regime in which Theorem~\ref{thm:water_fill} holds.}
\label{tab:closure}
\centering
\scriptsize
\setlength{\tabcolsep}{4pt}
\renewcommand{\arraystretch}{0.95}
\begin{tabular}{p{0.16\linewidth} *{4}{P{0.075\linewidth}} | *{4}{P{0.075\linewidth}}}
\toprule
& \multicolumn{4}{c|}{\textbf{RSNA-ICH}} & \multicolumn{4}{c}{\textbf{CT-RATE}} \\
\cmidrule(lr){2-5}\cmidrule(lr){6-9}
VLM & $\gamma_K$ & $1/\gamma_K$ & $\hat{r}$ & Closure & $\gamma_K$ & $1/\gamma_K$ & $\hat{r}$ & Closure \\
\midrule
MedGemma 1.5 4B & 0.279 & $3.58\times$ & $3.41\times$  & \textbf{4.9\%}     & 0.171 & $5.85\times$ & $4.21\times$  & \textbf{28.0\%} \\
MedGemma 4B     & 0.327 & $3.06\times$ & $4.30\times$  & \underline{40.6\%} & 0.200 & $5.00\times$ & $7.49\times$  & \underline{49.8\%} \\
MedGemma 27B    & 0.265 & $3.77\times$ & $4.85\times$  & \textbf{28.6\%}    & 0.180 & $5.56\times$ & $6.70\times$  & \textbf{20.6\%} \\
Hulu-Med 7B     & 0.206 & $4.85\times$ & $40.1\times$  & $727\%$            & 0.132 & $7.58\times$ & $38.0\times$  & $401\%$ \\
Hulu-Med 4B     & 0.253 & $3.95\times$ & $60.9\times$  & $1440\%$           & 0.162 & $6.17\times$ & $100.3\times$ & $1524\%$ \\
Hulu-Med 32B    & 0.224 & $4.46\times$ & $3.38\times$  & \textbf{24.2\%}    & 0.152 & $6.58\times$ & $3.72\times$  & \underline{43.5\%} \\
\bottomrule
\end{tabular}
\end{table*}

\subsection{Ablation Studies}
\label{sec:exp:ablation}

\noindent\textbf{Quantizer attribution.} We substitute Gain-shape SQ in the 3-axis DCT + reverse water-filling pipeline with two alternatives, a symmetric uniform quantizer with per-vector absmax scaling and NormalFloat~\cite{dettmers2023qlora}.  Table~\ref{tab:abl_quantizer} shows that gain-shape SQ achieves the lowest flip rate across all settings, with the largest gap at 1-bit. We attribute this gap to the heavy-tailed post-DCT coefficient distribution (Figure~\ref{fig:quant_mechanism}, median absmax-to-std $4.81$ versus the Gaussian reference $3.01$ at $d{=}256$). Naive absmax scaling assigns codebook levels to the spectrum tail rather than the bulk, while gain-shape SQ's decomposition factors out per-vector magnitude before quantization. Per-quantizer reconstruction MSE is in Appendix~\ref{app:quant_mechanism}.

\begin{table}[t]
\caption{\textbf{Quantizer attribution: macro flip rate (\%) on RSNA-ICH under three scalar quantizers inside the 3-axis DCT + reverse water-filling pipeline.} The DCT and water-filling stages are held fixed and only the per-bin scalar quantizer differs. Gain-shape SQ is the configuration used in TACT-KV. \textbf{Bold} marks the lowest entry in a column, \underline{underline} marks the second-lowest. Each entry aggregates $18{,}000$ binary predictions.}
\label{tab:abl_quantizer}
\centering
\scriptsize
\setlength{\tabcolsep}{4pt}
\renewcommand{\arraystretch}{1.05}
\begin{tabular}{l *{4}{P{0.06\linewidth}} | *{4}{P{0.06\linewidth}}}
\toprule
& \multicolumn{4}{c|}{\textbf{MedGemma 1.5 4B}} & \multicolumn{4}{c}{\textbf{Hulu-Med 7B}} \\
\cmidrule(lr){2-5}\cmidrule(lr){6-9}
Quantizer & 1-bit & 2-bit & 3-bit & 4-bit & 1-bit & 2-bit & 3-bit & 4-bit \\
\midrule
Gain-shape SQ      & \textbf{3.78}    & \textbf{1.91}    & \textbf{1.29}    & \textbf{1.09}    & \textbf{6.08}    & \textbf{3.51}    & \textbf{2.33}    & \textbf{1.86} \\
Uniform sym        &          24.04   &          15.56   &           8.45   &          1.41  & \underline{36.25}&          25.88 & \underline{4.94} & \underline{2.12} \\
NormalFloat~\cite{dettmers2023qlora}            & \underline{23.26}& \underline{12.40}& \underline{5.34} &\underline{1.29} &          37.05   & \underline{24.41}   &          11.74   &          6.60 \\
\bottomrule
\end{tabular}
\end{table}

\begin{wraptable}{r}{0.50\textwidth}
\vspace{-\intextsep}
\centering
\caption{\textbf{Transform and allocation ablations.} RSNA-ICH flip rate (\%) at 1 bit.}
\label{tab:abl_axis}
\scriptsize
\setlength{\tabcolsep}{3pt}
\renewcommand{\arraystretch}{1.05}
\begin{tabular}{@{}l *{2}{P{0.225\linewidth}}@{}}
\toprule
& \textbf{MedGemma} & \textbf{Hulu-Med} \\
Configuration & \textbf{1.5 4B} & \textbf{7B} \\
\midrule
3-axis DCT + water-filling (ours) & \textbf{3.78} & \textbf{6.08} \\
2-axis DCT + water-filling & \underline{5.00} & \underline{6.21} \\
1-axis DCT + water-filling & 6.00 & 9.65 \\
w/o DCT + water-filling & 6.61 & 42.20 \\
3-axis DCT + uniform allocation & 9.97 & 25.76 \\
\bottomrule
\end{tabular}
\end{wraptable}

\smallskip\noindent\textbf{Transform and bit-allocation ablations.}
With gain-shape SQ fixed, Table~\ref{tab:abl_axis} isolates the roles of the transform and bit allocation. The reduced-axis variants follow the axis choices of FreqKV~\cite{kai2025freqkv} and Fourier-VLM~\cite{wang2025fourier}. Flip rate increases as fewer spatial axes are transformed. Removing the DCT is particularly costly on Hulu-Med 7B, where flip rate rises from $6.08\%$ to $42.20\%$. Uniform allocation also degrades both models, showing that the transform and adaptive allocation each contribute to prediction fidelity.
\WFclear

\subsection{Cross-Domain Generalization: Video-MME}
\label{sec:exp:video}

We evaluate on Video-MME (no-subtitle) across four general-purpose VLMs. Table~\ref{tab:video_acc} reports per-question accuracy on the four-way MCQ with Wald $95\%$ CI half-widths. TACT-KV is the highest-accuracy method on every VLM at 1-bit, and the worst-case drop from bf16 is $4.77$\,pp on Qwen3-VL-8B. Other baselines collapse at 1-bit (KIVI below $13\%$ on every VLM). The advantage of three-axis structural compression therefore extends beyond medical imaging to video tasks.

\begin{table*}[t]
\caption{\textbf{Video-MME no-subtitle accuracy (\%) of the compressed cache across four general-purpose VLMs.} Each entry is per-question accuracy on $2{,}700$ four-way MCQs followed by the Wald $95\%$ CI half-width. \textbf{Bold} marks the highest accuracy in a column, \underline{underline} marks the second-highest, ranked over methods with measured values.}
\label{tab:video_acc}
\centering
\scriptsize
\setlength{\tabcolsep}{3pt}
\renewcommand{\arraystretch}{1.0}
\begin{tabular}{l *{4}{P{0.09\linewidth}} | *{4}{P{0.09\linewidth}}}
\toprule
Method & 1-bit & 2-bit & 3-bit & 4-bit & 1-bit & 2-bit & 3-bit & 4-bit \\
\midrule

\rowcolor{vlmgray} & \multicolumn{4}{c|}{\textbf{\textit{Qwen2.5-VL-7B (bf16: $60.00$\textsubscript{$\pm$1.85})}}} & \multicolumn{4}{c}{\textbf{\textit{Qwen3-VL-8B (bf16: $72.33$\textsubscript{$\pm$1.69})}}} \\
TACT-KV (ours) & \textbf{57.15}\textsubscript{$\pm$1.87} & 59.52\textsubscript{$\pm$1.85} & \underline{60.00}\textsubscript{$\pm$1.85} & \underline{59.93}\textsubscript{$\pm$1.85} & \textbf{67.56}\textsubscript{$\pm$1.77} & 71.00\textsubscript{$\pm$1.71} & \underline{71.69}\textsubscript{$\pm$1.70} & \underline{71.93}\textsubscript{$\pm$1.69} \\
TurboQuant     & 23.04\textsubscript{$\pm$1.59} & 25.04\textsubscript{$\pm$1.63} & 21.33\textsubscript{$\pm$1.55} & 44.85\textsubscript{$\pm$1.88} & 14.33\textsubscript{$\pm$1.32} & 54.26\textsubscript{$\pm$1.88} & 70.93\textsubscript{$\pm$1.71} & 71.67\textsubscript{$\pm$1.70} \\
KIVI           & 12.04\textsubscript{$\pm$1.23} & \underline{59.70}\textsubscript{$\pm$1.85} & \textbf{60.37}\textsubscript{$\pm$1.84} & \textbf{59.96}\textsubscript{$\pm$1.85} & 3.44\textsubscript{$\pm$0.69} & 71.30\textsubscript{$\pm$1.71} & \textbf{71.89}\textsubscript{$\pm$1.70} & \textbf{72.22}\textsubscript{$\pm$1.69} \\
VidKV          & 19.70\textsubscript{$\pm$1.50} & 59.33\textsubscript{$\pm$1.85} & -- & -- & \underline{26.63}\textsubscript{$\pm$1.67} & \textbf{71.67}\textsubscript{$\pm$1.70} & -- & -- \\
AKVQ-VL        & \underline{39.44}\textsubscript{$\pm$1.84} & \textbf{59.89}\textsubscript{$\pm$1.85} & -- & -- & 7.37\textsubscript{$\pm$0.99} & \underline{71.37}\textsubscript{$\pm$1.71} & -- & -- \\

\midrule

\rowcolor{vlmgray} & \multicolumn{4}{c|}{\textbf{\textit{LLaVA-OneVision-7B (bf16: $59.26$\textsubscript{$\pm$1.85})}}} & \multicolumn{4}{c}{\textbf{\textit{VideoLLaMA3-7B (bf16: $65.19$\textsubscript{$\pm$1.80})}}} \\
TACT-KV (ours) & \textbf{55.63}\textsubscript{$\pm$1.87} & \underline{58.70}\textsubscript{$\pm$1.86} & \textbf{58.78}\textsubscript{$\pm$1.86} & \underline{58.93}\textsubscript{$\pm$1.86} & \textbf{62.00}\textsubscript{$\pm$1.83} & \underline{64.89}\textsubscript{$\pm$1.80} & \textbf{65.07}\textsubscript{$\pm$1.80} & \textbf{64.93}\textsubscript{$\pm$1.80} \\
TurboQuant     & 39.56\textsubscript{$\pm$1.84} & 47.67\textsubscript{$\pm$1.88} & 40.22\textsubscript{$\pm$1.85} & 38.04\textsubscript{$\pm$1.83} & \underline{40.81}\textsubscript{$\pm$1.85} & 56.44\textsubscript{$\pm$1.87} & 27.41\textsubscript{$\pm$1.68} & 52.78\textsubscript{$\pm$1.88} \\
KIVI           & 7.63\textsubscript{$\pm$1.00} & 58.30\textsubscript{$\pm$1.86} & \underline{58.59}\textsubscript{$\pm$1.86} & \textbf{59.48}\textsubscript{$\pm$1.85} & 7.44\textsubscript{$\pm$0.99} & 64.67\textsubscript{$\pm$1.80} & \underline{65.04}\textsubscript{$\pm$1.80} & \underline{64.70}\textsubscript{$\pm$1.80} \\
VidKV          & \underline{40.00}\textsubscript{$\pm$1.85} & \textbf{58.85}\textsubscript{$\pm$1.86} & -- & -- & 40.56\textsubscript{$\pm$1.85} & \textbf{65.04}\textsubscript{$\pm$1.80} & -- & -- \\
AKVQ-VL        & 18.96\textsubscript{$\pm$1.48} & 58.56\textsubscript{$\pm$1.86} & -- & -- & 12.44\textsubscript{$\pm$1.25} & 64.19\textsubscript{$\pm$1.81} & -- & -- \\

\bottomrule
\end{tabular}
\end{table*}

\subsection{Deployment Profiling}
\label{sec:exp:deployment}

We profile the iso-HBM decoding behavior of TACT-KV on a single $96$~GiB NVIDIA RTX PRO 6000 GPU at $S = 60$ axial slices. To avoid materializing the full reconstructed cache at decode, we implement a tile-fused inverse path that integrates index dequantization and the inverse three-axis DCT into the FlashAttention~\cite{dao2022flashattention} kernel. The per-step intermediate footprint is then bounded by the FlashAttention block size and is independent of $B$ (kernel details in Appendix~\ref{app:kernel}).

\begin{table}[!ht]
\caption{\textbf{Memory and wall-clock performance on one 96 GiB RTX PRO 6000 GPU at $S=60$ slices.} Factors in parentheses give batch expansion, cache reduction, and throughput gain relative to bf16. Cache size is measured at $B=1$ after prefill. Decode step time and throughput are averaged over five CT-RATE volumes at each method's maximum batch. Headroom is the device-reported capacity (102.6 GB) minus model weights.}
\label{tab:deploy}
\centering
\scriptsize
\setlength{\tabcolsep}{3pt}
\renewcommand{\arraystretch}{1.0}
\begin{tabular}{l
    P{0.14\linewidth}<{\raggedright\arraybackslash}
    P{0.19\linewidth}<{\raggedright\arraybackslash}
    P{0.14\linewidth}<{\raggedright\arraybackslash}
    P{0.17\linewidth}<{\raggedright\arraybackslash}}
\toprule
Method & Max decoding $B$ & Per-request cache (MB) & Decode step (ms) & Throughput (tok/s) \\
\midrule
\rowcolor{vlmgray}\multicolumn{5}{c}{\textbf{\textit{Hulu-Med 4B}} (Weights 9.0 GB, HBM headroom 93.6 GB)} \\
bf16 & 32 & 2333 & 165 & 194 \\
TACT-KV & 128\textsubscript{$(4.00\times)$} & 309\textsubscript{$(7.55\times)$} & 467 & 274\textsubscript{$(1.42\times)$} \\
\rowcolor{vlmgray}\multicolumn{5}{c}{\textbf{\textit{Hulu-Med 7B}} (Weights 16.1 GB, HBM headroom 86.5 GB)} \\
bf16 & 64 & 915 & 136 & 472 \\
TACT-KV & 400\textsubscript{$(6.25\times)$} & 132\textsubscript{$(6.93\times)$} & 723 & 553\textsubscript{$(1.17\times)$} \\
\rowcolor{vlmgray}\multicolumn{5}{c}{\textbf{\textit{Hulu-Med 32B}} (Weights 66.4 GB, HBM headroom 36.2 GB)} \\
bf16 & 8 & 4140 & 121 & 66 \\
TACT-KV & 64\textsubscript{$(8.00\times)$} & 543\textsubscript{$(7.62\times)$} & 481 & 133\textsubscript{$(2.02\times)$} \\
\bottomrule
\end{tabular}
\end{table}

\noindent\textbf{Cache savings support larger batches and higher throughput.} We increase $B$ until the timed decode loop runs out of memory and report the largest batch that fits for each method in Table~\ref{tab:deploy}. Across the three scales, the per-request KV cache size shrinks by $6.9$--$7.6\times$, and the largest decoding batch a single GPU sustains grows by $4.00\times$, $6.25\times$, and $8.00\times$ respectively. The larger batches translate into $1.17$--$2.02\times$ higher aggregate decoding throughput. The batch ratio rises monotonically with the weight-to-headroom fraction. When weights occupy $66.4$~GB of $102.6$~GB and the cache becomes the dominant non-weight term (Hulu-Med 32B), the cache savings translate most directly into a batch-capacity multiplier. At smaller scales, the larger non-cache headroom dilutes the gain.

\begin{wraptable}{r}{0.50\textwidth}
\vspace{-\intextsep}
\centering
\caption{\textbf{Decode latency at matched batch sizes.} Each model uses bf16's maximum $B$. Parentheses show speedup.}
\label{tab:deploy_matched}
\scriptsize
\setlength{\tabcolsep}{3pt}
\renewcommand{\arraystretch}{1.05}
\begin{tabular}{@{}P{0.31\linewidth} P{0.10\linewidth} P{0.17\linewidth} P{0.31\linewidth}@{}}
\toprule
& & \multicolumn{2}{c}{Decode step (ms)} \\
\cmidrule(lr){3-4}
Model & $B$ & bf16 & TACT-KV \\
\midrule
Hulu-Med 4B & 32 & 165 & \textbf{114}\textsubscript{$(1.45\times)$} \\
Hulu-Med 7B & 64 & 136 & \textbf{119}\textsubscript{$(1.14\times)$} \\
Hulu-Med 32B & 8 & 121 & \textbf{110}\textsubscript{$(1.10\times)$} \\
\bottomrule
\end{tabular}
\end{wraptable}

\noindent\textbf{TACT-KV also reduces latency at the same batch size.} For each model, we use the largest batch supported by bf16 and retain the same five CT-RATE volumes. Table~\ref{tab:deploy_matched} shows that TACT-KV reduces step latency on all three models, with speedups of $1.10$--$1.45\times$. Both methods generate the same number of tokens per step at fixed $B$, so the latency reduction gives the same factor of improvement in aggregate throughput.

\WFclear

\section{Discussion and Conclusion}
We presented TACT-KV, a transform-coding pipeline for compressing the
visual KV cache of vision--language models on volumetric token grids.
TACT-KV reshapes the cache into a $(S, H, W)$ tensor, applies a
separable three-axis DCT-II for spatial decorrelation, allocates a
per-layer bit budget by reverse water-filling, and quantizes each
frequency bin with a scalar codebook. The pipeline is calibration-free
and adds no learned parameters. The volumetric prior we exploit is
absent from the text and natural-image settings that prior KV
compressors target, and we prove that the distortion gain of reverse
water-filling over uniform allocation has a closed form, $1/\gamma$,
in the spectral flatness of the transformed cache. Across six medical VLMs on two 3D CT benchmarks, TACT-KV achieves the lowest 1-bit flip rate on eleven of twelve model and dataset pairs in our main comparison. The method also transfers to four general-purpose VLMs on Video-MME, and its gain in maximum decoding batch size at equal HBM increases with model scale, reaching $8\times$ on Hulu-Med 32B.
The margin between TACT-KV and 1D quantizers on
these benchmarks also measures how much volumetric structure scalar
codebooks alone cannot recover, and that margin widens as the input
becomes more spatially coherent.
Integrating TACT-KV with attention-aware eviction on the text portion
of the cache is a natural next step and orthogonal to the
bit-allocation savings reported here.

\smallskip\noindent\textbf{Limitation.}
The closed-form prediction of Theorem~\ref{thm:water_fill} holds in
the high-rate regime in which every frequency bin receives
non-trivial bit allocation, and the lowest-rate operating points on
the smaller Hulu-Med variants fall outside this regime
(Section~\ref{sec:exp:closure}). Our pipeline also compresses the
visual portion of the cache only, leaving text tokens at bf16, so the
savings reported here are tied to inputs in which visual tokens
dominate. The volumetric prior is validated on chest CT and head CT
only, and whether it transfers to other 3D modalities such as MRI,
PET, and 3D ultrasound is not measured. Although the same pipeline
transfers to Video-MME without modification, the temporal axis differs
in origin from the slice axis, and the gains we observe there are
empirical rather than predicted by the volumetric assumption that
motivates the design.

\smallskip\noindent\textbf{Broader Impact.}
Lower memory requirements can broaden access to volumetric medical VLMs at clinical sites with limited GPU capacity~\cite{lan2026performance}. Prediction changes introduced by compression could lead to missed findings or unnecessary follow-up in clinical use. Evaluation on local data should therefore examine per-label prediction changes and diagnostic accuracy, especially for rare findings that aggregate metrics may obscure.

\section*{Acknowledgments}
This research was supported by the National Institutes of Health (NIH) under grant R01DK135597, with \$20,000 representing 66\% of the total project costs. The content is solely the responsibility of the authors and does not necessarily represent the official views of the National Institutes of Health. This material is based upon work supported by the U.S. National Science Foundation under Award Number 2434229. A total of \$5,000, representing 17\% of the total project costs, was financed with this funding. Any opinions, findings, and conclusions or recommendations expressed in this material are those of the author(s) and do not necessarily reflect the views of the U.S. National Science Foundation. Non-governmental support was provided by the Vanderbilt Internal Grants. A total of \$5,000, representing 17\% of the total project costs, was financed with non-governmental funds.

The authors declare no additional competing interests.





\bibliography{refs}
\bibliographystyle{plain}

\newpage
\clearpage
\appendix

\renewcommand{\thepart}{}
\renewcommand{\partname}{}

\part{Appendix}
\parttoc

\clearpage

\section{Proof of Theorem~\ref{thm:water_fill}}
\label{app:proof}

We give the full proof of Theorem~\ref{thm:water_fill} from Section~\ref{sec:method:waterfill}. The proof proceeds under the continuous-bit relaxation of~\eqref{eq:waterfill} and the high-rate Gaussian distortion model used to derive the per-bin distortion $\mathcal{D}_i(b_i) = c \sigma_i^2 \cdot 2^{-2 b_i}$. The two regime caveats noted in the footnote of Section~\ref{sec:method:waterfill}---integer rounding of $b_i$ and breakdown of the high-rate model at very low average rate---are discussed in Remarks~\ref{rem:integer} and~\ref{rem:lowrate} at the end of this appendix.

\subsection*{Preliminaries}

Fix a layer index $\ell$ and a per-bin variance vector $(\sigma_1^2, \ldots, \sigma_N^2)$ with $\sigma_i^2 > 0$, defined as in~\eqref{eq:variance} of Section~\ref{sec:method:waterfill}. The high-rate Gaussian scalar-quantization model assigns to each bin the per-bin distortion
\begin{equation}
\label{eq:proof_distortion_model}
\mathcal{D}_i(b_i) = c \cdot \sigma_i^2 \cdot 2^{-2 b_i}, \qquad i = 1, \ldots, N,
\end{equation}
where $c > 0$ is the quantizer-geometry constant (independent of $i$). Let $\mathcal{D}(\mathbf{b}) = \sum_{i=1}^N \mathcal{D}_i(b_i)$ denote the total reconstruction MSE under bit allocation $\mathbf{b} = (b_1, \ldots, b_N)$. We work under the continuous relaxation $b_i \in \mathbb{R}_{\geq 0}$ and the matched budget constraint $\sum_{i=1}^N b_i = R_\ell$.

For convenience we restate the theorem before proving it.

\setcounter{theorem}{0}
\begin{theorem}[restated from Section~\ref{sec:method:waterfill}]
Let $\{\sigma_i^2\}_{i=1}^N$ be per-bin variances, and define the spectral flatness
\begin{equation*}
\gamma = \frac{\left(\prod_{i=1}^N \sigma_i^2\right)^{1/N}}{(1/N) \sum_{i=1}^N \sigma_i^2} \in (0, 1].
\end{equation*}
Under matched bit budget $R_\ell$ and the high-rate Gaussian distortion model~\eqref{eq:proof_distortion_model}, the total distortion under uniform allocation and under reverse water-filling allocation satisfies
\begin{equation*}
\frac{\mathcal{D}_{\mathrm{uniform}}}{\mathcal{D}_{\mathrm{water\text{-}fill}}} = \frac{1}{\gamma}.
\end{equation*}
\end{theorem}

\begin{proof}
The proof has five steps. Step~1 evaluates the uniform-allocation distortion. Steps~2 and~3 derive the water-filling solution from the KKT conditions of the constrained optimization. Step~4 specializes to the high-rate regime in which all bins are active. Step~5 forms the ratio.

\medskip
\noindent\textbf{Step 1: Uniform-allocation distortion.}
Setting $b_i = R_\ell / N$ for every $i$ in~\eqref{eq:proof_distortion_model} gives
\begin{equation}
\label{eq:proof_uniform}
\mathcal{D}_{\mathrm{uniform}} \;=\; \sum_{i=1}^N c \cdot \sigma_i^2 \cdot 2^{-2 R_\ell / N} \;=\; c \cdot 2^{-2 R_\ell / N} \cdot \sum_{i=1}^N \sigma_i^2.
\end{equation}

\medskip
\noindent\textbf{Step 2: Convexity and KKT conditions for water-filling.}
The water-filling problem is the constrained optimization
\begin{equation}
\label{eq:proof_opt}
\min_{\mathbf{b} \in \mathbb{R}^N_{\geq 0}} \;\; \sum_{i=1}^N c \cdot \sigma_i^2 \cdot 2^{-2 b_i} \quad \text{subject to} \quad \sum_{i=1}^N b_i \;=\; R_\ell.
\end{equation}
The objective is a sum of strictly convex functions of the form $b \mapsto c \sigma_i^2 \cdot 2^{-2 b}$ (the second derivative is $4 (\ln 2)^2 \cdot c \sigma_i^2 \cdot 2^{-2b} > 0$). The equality constraint is affine and the inequality constraints $b_i \geq 0$ are convex, so~\eqref{eq:proof_opt} is a convex program. The KKT conditions are therefore both necessary and sufficient for global optimality.

Introduce a multiplier $\mu \in \mathbb{R}$ for the budget equality and multipliers $\nu_i \geq 0$ for the constraints $-b_i \leq 0$. The Lagrangian is
\begin{equation*}
L(\mathbf{b}, \mu, \boldsymbol{\nu}) \;=\; \sum_{i=1}^N c \sigma_i^2 \cdot 2^{-2 b_i} \;+\; \mu \!\left(\sum_{i=1}^N b_i - R_\ell\right) \;-\; \sum_{i=1}^N \nu_i b_i.
\end{equation*}
Stationarity in $b_i$ yields
\begin{equation}
\label{eq:proof_stationarity}
\frac{\partial L}{\partial b_i} \;=\; -2 \ln 2 \cdot c \sigma_i^2 \cdot 2^{-2 b_i} \;+\; \mu \;-\; \nu_i \;=\; 0
\end{equation}
for every $i$. Define the water level
\begin{equation}
\label{eq:proof_lambda_def}
\lambda \;:=\; \frac{\mu}{2 \, c \, \ln 2},
\end{equation}
which is positive because $c > 0$ and $\mu > 0$ (any candidate optimum has $\mu > 0$ since the objective strictly decreases in each $b_i$). Equation~\eqref{eq:proof_stationarity} rearranges to $c \sigma_i^2 \cdot 2^{-2 b_i} = (\mu - \nu_i) / (2 \ln 2)$, equivalently
\begin{equation}
\label{eq:proof_per_bin}
\sigma_i^2 \cdot 2^{-2 b_i} \;=\; \lambda - \frac{\nu_i}{2 \, c \, \ln 2}.
\end{equation}

\medskip
\noindent\textbf{Step 3: Active set and per-bin distortion.}
Complementary slackness gives $\nu_i \cdot b_i = 0$ for each $i$. We split bins into two cases.

\emph{Active bins.} If $b_i > 0$, then $\nu_i = 0$ and~\eqref{eq:proof_per_bin} reduces to $\sigma_i^2 \cdot 2^{-2 b_i} = \lambda$, equivalent to
\begin{equation}
\label{eq:proof_active_bits}
b_i \;=\; \tfrac{1}{2} \log_2(\sigma_i^2 / \lambda).
\end{equation}
For~\eqref{eq:proof_active_bits} to give $b_i > 0$ we need $\sigma_i^2 > \lambda$. The corresponding per-bin distortion under~\eqref{eq:proof_distortion_model} is
\begin{equation}
\label{eq:proof_active_distortion}
\mathcal{D}_i \;=\; c \cdot \sigma_i^2 \cdot 2^{-2 b_i} \;=\; c \, \lambda,
\end{equation}
which is the same constant for every active bin.

\emph{Inactive bins.} If $b_i = 0$, then~\eqref{eq:proof_stationarity} gives $\nu_i = \mu - 2 \, c \sigma_i^2 \, \ln 2 = 2 \, c \, \ln 2 \, (\lambda - \sigma_i^2)$, and the constraint $\nu_i \geq 0$ requires $\sigma_i^2 \leq \lambda$.

Defining the active set
\begin{equation*}
A \;:=\; \{ i : \sigma_i^2 > \lambda \}, \qquad K := |A|,
\end{equation*}
we have $b_i > 0$ if and only if $i \in A$, and the per-active-bin distortion is the constant $c\lambda$.

\medskip
\noindent\textbf{Step 4: High-rate regime.}
The high-rate Gaussian distortion model presumes that every bin receives non-trivial bit allocation, i.e., that $R_\ell$ is large enough that the optimal water level $\lambda$ falls strictly below the smallest variance:
\begin{equation}
\label{eq:proof_highrate}
\lambda \;<\; \min_{i} \sigma_i^2.
\end{equation}
In this regime $A = \{1, \ldots, N\}$ and $K = N$, so~\eqref{eq:proof_active_bits} and~\eqref{eq:proof_active_distortion} hold for every $i$. Substituting the per-bin allocation~\eqref{eq:proof_active_bits} into the budget constraint $\sum_{i=1}^N b_i = R_\ell$ yields
\begin{equation*}
\sum_{i=1}^N \tfrac{1}{2} \log_2(\sigma_i^2 / \lambda) \;=\; R_\ell
\;\;\Longleftrightarrow\;\;
\tfrac{1}{2}\!\left(\sum_{i=1}^N \log_2 \sigma_i^2 \;-\; N \log_2 \lambda \right) \;=\; R_\ell,
\end{equation*}
which solves to
\begin{equation}
\label{eq:proof_lambda_solved}
\log_2 \lambda \;=\; \frac{1}{N} \sum_{i=1}^N \log_2 \sigma_i^2 \;-\; \frac{2 R_\ell}{N}, \qquad \text{equivalently} \qquad \lambda \;=\; \left(\prod_{i=1}^N \sigma_i^2\right)^{1/N} \cdot 2^{-2 R_\ell / N}.
\end{equation}
The total water-filling distortion is therefore
\begin{equation}
\label{eq:proof_water_total}
\mathcal{D}_{\mathrm{water\text{-}fill}} \;=\; \sum_{i=1}^N c \, \lambda \;=\; c \, N \, \lambda \;=\; c \, N \, \left(\prod_{i=1}^N \sigma_i^2\right)^{1/N} \cdot 2^{-2 R_\ell / N}.
\end{equation}

\medskip
\noindent\textbf{Step 5: Ratio.}
Dividing~\eqref{eq:proof_uniform} by~\eqref{eq:proof_water_total},
\begin{equation*}
\frac{\mathcal{D}_{\mathrm{uniform}}}{\mathcal{D}_{\mathrm{water\text{-}fill}}}
\;=\;
\frac{c \cdot 2^{-2 R_\ell / N} \cdot \sum_{i=1}^N \sigma_i^2}
     {c \, N \, \left(\prod_{i=1}^N \sigma_i^2\right)^{1/N} \cdot 2^{-2 R_\ell / N}}
\;=\;
\frac{(1/N) \sum_{i=1}^N \sigma_i^2}{\left(\prod_{i=1}^N \sigma_i^2\right)^{1/N}}
\;=\;
\frac{1}{\gamma},
\end{equation*}
where the last equality uses the definition of $\gamma$ in~\eqref{eq:flatness}. Both occurrences of $c$ and the common factor $2^{-2 R_\ell / N}$ cancel, so the ratio is independent of the bit budget and of the quantizer-geometry constant. \qedhere
\end{proof}

\subsection*{Remarks}

\begin{remark}[Range of $\gamma$ and tightness]\label{rem:gamma_range}
The arithmetic-geometric mean inequality applied to the positive reals $\sigma_1^2, \ldots, \sigma_N^2$ gives
\begin{equation*}
\frac{1}{N} \sum_{i=1}^N \sigma_i^2 \;\geq\; \left(\prod_{i=1}^N \sigma_i^2\right)^{1/N},
\end{equation*}
with equality if and only if $\sigma_1^2 = \cdots = \sigma_N^2$. Combined with $\sigma_i^2 > 0$ for all $i$, this shows $\gamma \in (0, 1]$, and consequently $1/\gamma \geq 1$. The water-filling advantage equals one (uniform is already optimal) precisely when the cache spectrum is flat, and grows without bound as the spectrum concentrates on a small number of bins.
\end{remark}

\begin{remark}[Integer rounding penalty]\label{rem:integer}
The implementation rounds $b_i$ via the floor in~\eqref{eq:waterfill} and applies a greedy $\pm 1$ correction to enforce $\sum_i b_i = R_\ell$ exactly. This restricts the feasible set of bit allocations to $\mathbb{Z}_{\geq 0}^N$, so the integer-allocated water-fill distortion satisfies
\begin{equation*}
\mathcal{D}^{\mathrm{int}}_{\mathrm{water\text{-}fill}} \;\geq\; \mathcal{D}^{\mathrm{cont}}_{\mathrm{water\text{-}fill}},
\end{equation*}
where the right-hand side is the continuous-relaxation value used in the proof above. The empirical ratio $\mathcal{D}_{\mathrm{uniform}} / \mathcal{D}^{\mathrm{int}}_{\mathrm{water\text{-}fill}}$ is therefore tightened toward $1$ relative to the closed-form $1/\gamma$, and is visible as a small finite-rate gap on the points inside the closure band of Figure~\ref{fig:closure}.
\end{remark}

\begin{remark}[Breakdown of the high-rate model at low average rate]\label{rem:lowrate}
The high-rate Gaussian distortion model~\eqref{eq:proof_distortion_model} is an asymptotic approximation for large $b_i$ and Gaussian-distributed source coordinates. When the average bit rate $R_\ell / N$ becomes small, two effects pull the empirical ratio away from the closed-form $1/\gamma$. First, condition~\eqref{eq:proof_highrate} can fail and bins with $\sigma_i^2 \leq \lambda$ receive zero bits, in which case the active-set argument in Step~3 yields a partition-dependent expression for $\mathcal{D}_{\mathrm{water\text{-}fill}}$ rather than the clean $cN\lambda$ form of Step~4. Second, on heavy-tailed empirical post-DCT distributions (most pronounced for the Hulu-Med 4B/7B K-cache in Section~\ref{sec:exp:closure}), the actual scalar-quantization distortion at $b_i = 1$ deviates from the high-rate prediction $c \sigma_i^2 \cdot 2^{-2}$ in a direction that disproportionately penalizes uniform allocation, since uniform spreads bits across low-energy bins where the high-rate model is loosest. The two effects can combine, and the proof above does not control either, which is why we report the four out-of-band entries in Section~\ref{sec:exp:closure} for completeness rather than as confirmations of the closed form.
\end{remark}

\section{Per-Axis Spectral Flatness}
\label{app:per_axis_gamma}

This appendix reports the per-axis spectral flatness of the visual KV cache that motivates the 3-axis design.

\subsection*{Protocol}

For each VLM-dataset pair we sample $20$ studies, run prefill at the per-VLM official slice budget, and read out the per-bin variance tensor $\sigma^2 \in \mathbb{R}_{\geq 0}^{S \times H \times W}$ defined in~\eqref{eq:variance}. For each axis $\alpha \in \{s, h, w\}$ with corresponding size $N_\alpha \in \{S, H, W\}$ we project $\sigma^2$ onto its 1-D marginal by summing over the other two axes,
\begin{equation*}
m_s[i] = \sum_{j=1}^{H} \sum_{k=1}^{W} \sigma^2_{ijk}, \qquad m_h[j] = \sum_{i=1}^{S} \sum_{k=1}^{W} \sigma^2_{ijk}, \qquad m_w[k] = \sum_{i=1}^{S} \sum_{j=1}^{H} \sigma^2_{ijk},
\end{equation*}
and apply the flatness functional of~\eqref{eq:flatness} to the marginal,
\begin{equation*}
\gamma_K^\alpha = \frac{\big(\prod_{i=1}^{N_\alpha} m_\alpha[i]\big)^{1/N_\alpha}}{(1/N_\alpha) \sum_{i=1}^{N_\alpha} m_\alpha[i]} \in (0, 1].
\end{equation*}
We report per-axis $\gamma_K^\alpha$ averaged across all transformer layers and all sampled studies for each pair.

\subsection*{Results}

Table~\ref{tab:per_axis_gamma_K} reports $\gamma_K^\alpha$ on the K cache across the six medical VLMs and two CT benchmarks.

\begin{table*}[t]
\caption{\textbf{Per-axis spectral flatness $\gamma_K^\alpha$ of the visual K cache.} The joint $\gamma_K$ column reproduces the value reported in Table~\ref{tab:closure}. Per-axis $\gamma_K^\alpha$ is computed on the marginal variance projection along axis $\alpha$, averaged over transformer layers and over $20$ sampled studies per pair.}
\label{tab:per_axis_gamma_K}
\centering
\scriptsize
\setlength{\tabcolsep}{4pt}
\renewcommand{\arraystretch}{0.95}
\begin{tabular}{p{0.16\linewidth} *{4}{P{0.075\linewidth}} | *{4}{P{0.075\linewidth}}}
\toprule
& \multicolumn{4}{c|}{\textbf{RSNA-ICH}} & \multicolumn{4}{c}{\textbf{CT-RATE}} \\
\cmidrule(lr){2-5}\cmidrule(lr){6-9}
VLM & $\gamma_K$ & $\gamma_K^s$ & $\gamma_K^h$ & $\gamma_K^w$ & $\gamma_K$ & $\gamma_K^s$ & $\gamma_K^h$ & $\gamma_K^w$ \\
\midrule
MedGemma 1.5 4B & 0.279 & 0.389 & 0.458 & 0.412 & 0.171 & 0.215 & 0.473 & 0.447 \\
MedGemma 4B     & 0.327 & 0.448 & 0.534 & 0.468 & 0.200 & 0.249 & 0.534 & 0.480 \\
MedGemma 27B    & 0.265 & 0.443 & 0.440 & 0.379 & 0.180 & 0.258 & 0.432 & 0.396 \\
Hulu-Med 7B     & 0.206 & 0.416 & 0.354 & 0.273 & 0.132 & 0.279 & 0.345 & 0.271 \\
Hulu-Med 4B     & 0.253 & 0.444 & 0.392 & 0.336 & 0.162 & 0.282 & 0.367 & 0.320 \\
Hulu-Med 32B    & 0.224 & 0.500 & 0.415 & 0.300 & 0.152 & 0.350 & 0.403 & 0.296 \\
\bottomrule
\end{tabular}
\end{table*}

\noindent\textbf{Every axis is spectrally concentrated.} The largest entry of $\gamma_K^\alpha$ across the $36$ axis-VLM-dataset entries of Table~\ref{tab:per_axis_gamma_K} is $0.534$, on MedGemma 4B at the H axis on both datasets, so $\gamma_K^\alpha < 0.55$ everywhere. The marginal-wise water-filling advantage from Theorem~\ref{thm:water_fill} therefore satisfies $1/\gamma_K^\alpha > 1.87\times$ on every axis. Each dropped axis forfeits at least this margin under the marginal-wise application of the theorem, in agreement with the monotone $3 \to 2 \to 1$-axis degradation in Table~\ref{tab:abl_axis}. The pattern of axis-wise minima is split between the slice axis on six pairs and the width axis on the other six, with the H axis never the most concentrated, so any single dropped axis includes a strongly concentrated direction on at least half of the pairs.

\section{Reproducibility Statement}
\label{app:reproducibility}
\subsection{Dataset Details}
\label{app:dataset}

This appendix documents the official source, license, and sampling protocol of each benchmark used in Section~\ref{sec:experiments}, in the order they appear in Section~\ref{sec:exp:setup}.

\medskip
\noindent\textbf{RSNA-ICH.} The 2019 RSNA Intracranial Hemorrhage Detection Challenge releases a public head-CT dataset under the RSNA Brain CT Hemorrhage Challenge Terms of Use and Attribution~\cite{flanders2020construction}. Access is granted for non-commercial academic research and education, redistribution of any portion of the data is prohibited, and any use must cite the dataset as ``Brain CT Hemorrhage Dataset, Copyright RSNA, 2019'' alongside the Flanders et al.\ reference. Each study carries six binary labels, the global \texttt{any} flag and the five hemorrhage subtypes \texttt{epidural}, \texttt{intraparenchymal}, \texttt{intraventricular}, \texttt{subarachnoid}, and \texttt{subdural}. After de-duplication and removal of studies with corrupted axial DICOM windows, $N = 21{,}744$ usable training-split studies remain. We evaluate on a stratified subset of $N = 3{,}000$ studies obtained by stratifying the full pool on the (\texttt{any}, \texttt{epidural}) cross-product, the rare-subtype \texttt{epidural} drives the smallest stratum and is locked in to remove sample-to-sample variance across methods, then proportionally allocating $3{,}000$ slots and drawing without replacement at fixed seed $20260423$. Table~\ref{tab:rsna_distribution} reports the per-label prevalence in the resulting subset alongside the full training pool. Every label prevalence in the subset matches the training pool to within $0.23$ percentage points, so the subset preserves the underlying class distribution rather than selecting a sweet spot. The sampling script and the resulting manifest of study UIDs are released with the supplementary code.

\begin{table}[t]
\caption{\textbf{Per-label prevalence on the RSNA-ICH evaluation subset versus the full training pool.} The subset is obtained by proportional stratified sampling on the (\texttt{any}, \texttt{epidural}) cross-product at seed $20260423$. Differences in prevalence are reported in absolute percentage points.}
\label{tab:rsna_distribution}
\centering
\scriptsize
\setlength{\tabcolsep}{6pt}
\renewcommand{\arraystretch}{1.0}
\begin{tabular}{l P{0.10\linewidth} P{0.10\linewidth} | P{0.10\linewidth} P{0.10\linewidth} | P{0.08\linewidth}}
\toprule
& \multicolumn{2}{c|}{\textbf{Training pool} ($N{=}21{,}744$)} & \multicolumn{2}{c|}{\textbf{Subset} ($N{=}3{,}000$)} & \\
\cmidrule(lr){2-3}\cmidrule(lr){4-5}
Label & positives & rate (\%) & positives & rate (\%) & $\Delta$ (pp) \\
\midrule
\texttt{any}              & $8{,}882$ & $40.85$ & $1{,}226$ & $40.87$ & $+0.02$ \\
\texttt{epidural}         &     $354$ & $1.63$  &      $49$ & $1.63$  & $0.00$ \\
\texttt{intraparenchymal} & $5{,}321$ & $24.47$ &     $729$ & $24.30$ & $-0.17$ \\
\texttt{intraventricular} & $3{,}692$ & $16.98$ &     $506$ & $16.87$ & $-0.11$ \\
\texttt{subarachnoid}     & $3{,}932$ & $18.08$ &     $536$ & $17.87$ & $-0.21$ \\
\texttt{subdural}         & $3{,}812$ & $17.53$ &     $519$ & $17.30$ & $-0.23$ \\
\bottomrule
\end{tabular}
\end{table}

\medskip
\noindent\textbf{CT-RATE.} CT-RATE~\cite{hamamci2026generalist} is a public chest-CT benchmark released under CC BY-NC-SA 4.0, providing volumes paired with reconstructed reports and 18 binary multi-label abnormality annotations. The license restricts use to non-commercial academic research and prohibits redistribution of the dataset or any portion thereof. The official validation split contains $N = 3{,}039$ volumes from $1{,}304$ held-out patients. We evaluate on the full validation split with no further sampling. Each volume is processed under the per-VLM official preprocessing recipe (slice count and HU windowing inherited from the published evaluation pipeline of each VLM family), and predictions are made on each of the $18$ abnormalities independently for every volume, yielding $18 \times 3{,}039 = 54{,}702$ binary predictions per entry of the main results table.

\medskip
\noindent\textbf{Video-MME.} Video-MME~\cite{fu2025video} is a multimodal benchmark covering $900$ videos and $2{,}700$ human-annotated four-way multiple-choice questions across short, medium, and long durations. The official terms restrict use to academic research and prohibit commercial use, distribution, publication, copying, dissemination, or modification of the data without prior approval from the authors. We evaluate on all $2{,}700$ questions in the no-subtitle setting using the official Video-MME no-subtitle prompt verbatim, and extract the answer letter via the rule-based regex of the \texttt{lmms-eval} reference implementation. Each VLM uses its own video sampling recipe drawn from the public reference for that model, with the per-VLM frame schedule, pixel budget, and processor settings documented in Appendix~\ref{app:impl}.

\subsection{Per-VLM Implementation Details}
\label{app:impl}

This appendix records the per-VLM configuration of every evaluation in Section~\ref{sec:experiments}, in the order each topic is referenced from the main text.

\medskip
\noindent\textbf{Hardware, software stack, and compute resources.} All experiments run on a single $96$\,GiB NVIDIA RTX PRO 6000 Blackwell Server Edition GPU. Models are loaded in \texttt{bfloat16} via the HuggingFace \texttt{transformers} library with \texttt{attn\_implementation="flash\_attention\_2"}. The KV-cache compression backends are PyTorch reimplementations integrated through the standard \texttt{transformers.cache\_utils.DynamicCache} API and shipped with the supplementary code release. A single $\langle$VLM, dataset, method, bit$\rangle$ entry in Table~\ref{tab:flip_medical} or Table~\ref{tab:video_acc} corresponds to one full eval pass on this GPU, and the wall-clock per entry scales primarily with VLM size, slice budget, and the number of full-attention layers covered by the compression backend. For TACT-KV at $1$-bit, the RSNA-ICH pass ($N{=}3{,}000$ studies, $60$-slice budget) takes approximately $1.5$\,h on MedGemma 1.5 4B and MedGemma 4B (five full-attention layers compressed), $2.5$--$3$\,h on Hulu-Med 4B and Hulu-Med 7B (all transformer layers compressed), $3$\,h on MedGemma 27B, and $8$\,h on Hulu-Med 32B; the corresponding CT-RATE pass ($N{=}3{,}039$, $85$-slice budget) takes approximately $3.5$\,h on the 4B-scale variants, $3$\,h on Hulu-Med 7B, and $8$\,h on MedGemma 27B and Hulu-Med 32B. The Video-MME pass ($N{=}2{,}700$ four-way MCQs) finishes in under $3$\,h on Qwen2.5-VL-7B, LLaVA-OneVision-7B, and VideoLLaMA3-7B under their published sparse-sampling configurations, and in approximately a day on Qwen3-VL-8B under the dense cookbook configuration described below. Per-entry wall-clock for the four baseline backends sits within the same range, with the no-transform KIVI and TurboQuant somewhat faster than TACT-KV and the saliency-based AKVQ-VL somewhat slower at $1$-bit. Aggregated over the cells reported in the main tables, the ablations of Section~\ref{sec:exp:ablation}, and the Theorem~1 closure replicas of Section~\ref{sec:exp:closure}, the measured runs together account for on the order of $1{,}500$ GPU-hours on the GPU class above; preliminary kernel iterations, calibration sweeps, and runs discarded for protocol changes contribute a comparable additional budget, for a project total in the low thousands of GPU-hours.

\medskip
\noindent\textbf{Medical CT shared protocol.} For both RSNA-ICH and CT-RATE we use rank classification on assertion-style completion pairs. For each volume we build a single shared visual prefix that interleaves vision tokens with the text markers \texttt{SLICE 1}, \texttt{SLICE 2}, $\ldots$, prefill it once, snapshot the resulting cache, and for each abnormality forward only the completion tokens against the restored cache. The label is decided by comparing the mean per-token log-probability of the positive completion against the negative completion. RSNA-ICH uses the pair (\texttt{X hemorrhage is present.}, \texttt{X hemorrhage is not present.}) for $X \in \{$\texttt{intracranial}, \texttt{epidural}, \texttt{intraparenchymal}, \texttt{intraventricular}, \texttt{subarachnoid}, \texttt{subdural}$\}$. CT-RATE uses (\texttt{X is present.}, \texttt{X is not present.}) for each of the $18$ abnormalities. The shared-prefix cache pattern is reused identically across compression backends so that the evaluated quantization step is the only factor that changes between methods.

\medskip
\noindent\textbf{Slice budget and HU windowing.} Each volume is subsampled to a fixed slice budget along the axial axis using the MedGemma evenly-spaced indexing rule, $\text{idx}(i) = \lfloor i \cdot (n_{\mathrm{native}} - 1) / n_{\mathrm{budget}} \rceil$ for $i = 1, \ldots, n_{\mathrm{budget}}$. The default budget is $60$ slices for RSNA-ICH and $85$ slices for CT-RATE, matching the published MedGemma evaluation recipe. Each subsampled slice is mapped to a three-channel image by triple HU windowing with the MedGemma standard windows (R, G, B) $= ((-1024, 1024), (-135, 215), (0, 80))$. The same preprocessing is applied for all six medical VLMs to keep the visual input identical across methods and bit budgets.

\medskip
\noindent\textbf{Video-MME shared protocol.} For Video-MME we use the official no-subtitle prompt verbatim, \texttt{"Select the best answer to the following multiple-choice question based on the video. Respond with only the letter (A, B, C, or D) of the correct option.\textbackslash n\{question\}\textbackslash n\{options\}\textbackslash nThe best answer is:"}, with options as the four letter-prefixed strings shipped by the dataset. We generate up to $8$ tokens by greedy decoding (\texttt{do\_sample=False}) and extract the four-way letter via the rule-based regex of the \texttt{lmms-eval} reference implementation. The same prompt is shared by all four general-purpose VLMs.

\medskip
\noindent\textbf{MedGemma family (1.5 4B, 4B, 27B).} The MedGemma family is loaded from the official HuggingFace checkpoints \texttt{medgemma-1.5-4b-it}, \texttt{medgemma-4b-it}, and \texttt{medgemma-27b-it}. All three models inherit the Gemma-3 decoder, which interleaves five sliding-window layers per global full-attention layer. MedGemma 1.5 4B and MedGemma 4B each have $5$ full-attention layers and $29$ sliding-window layers ($34$ total). MedGemma 27B has $10$ full-attention and $52$ sliding-window layers ($62$ total). All three models set the sliding-window length to $1{,}024$ tokens, which already caps the per-layer cache on sliding-window layers via the standard dynamic sliding-window cache layer of the HuggingFace transformers library. We therefore leave sliding-window layers at bf16 and apply the compression backend only to full-attention layers, following the production convention of vLLM. The bypass is implemented uniformly across compression backends in the supplementary code release.

\medskip
\noindent\textbf{Hulu-Med family (4B, 7B, 32B).} The Hulu-Med family is loaded from the official HuggingFace checkpoints \texttt{Hulu-Med-4B}, \texttt{Hulu-Med-7B}, and \texttt{Hulu-Med-32B}. The family ships a custom processor and chat template that emit an image-token marker for each image block, so we load the weights through the standard causal-language-model loader with remote code enabled. Hulu-Med 4B uses a Qwen3 backbone with $36$ transformer layers, Hulu-Med 7B uses a Qwen2 backbone with $28$ layers, and Hulu-Med 32B uses a Qwen2 backbone with $64$ layers. The text-config of all three models marks every layer as full-attention and leaves the sliding-window field unset, so the compression backend applies to every transformer layer with no bypass. The image processor caps the total visual-token budget at approximately $16{,}000$ tokens and redistributes the per-slice token count when the input exceeds this budget, so the effective context length plateaus around $14$--$17$\,k tokens at slice budgets at or above $60$.

\medskip
\noindent\textbf{Qwen2.5-VL-7B-Instruct.} Loaded from the official HuggingFace checkpoint \texttt{Qwen2.5-VL}
\texttt{-7B-Instruct}. Video frames are sampled by the official Qwen-VL utility under the model-card sparse configuration, with a maximum pixel count of $360 \times 420$, fps $= 1.0$, and a frame cap of $32$. The dense cookbook configuration shipped with the Qwen3-VL repository drops Video-MME accuracy on the short-bucket by approximately $10$ percentage points on this model, so we follow the model card.

\medskip
\noindent\textbf{Qwen3-VL-8B-Instruct.} Loaded from the official HuggingFace checkpoint \texttt{Qwen3-VL-8B}
\texttt{-Instruct}. Video frames are sampled by the Qwen-VL utility under the dense cookbook configuration of the Qwen3-VL repository, with a total pixel budget of $20$\,M, a minimum pixel budget of $64 \cdot 32 \cdot 32$, sample fps $= 2$, a frame cap of $2{,}048$, image patch size $16$, and per-video metadata returned by the processor. The generic \texttt{lmms-eval} sparse configuration under-performs the cookbook by approximately $20$ percentage points on this model, so we follow the cookbook.

\medskip
\noindent\textbf{LLaVA-OneVision-7B.} Loaded from the official HuggingFace checkpoint \texttt{llava-onevision}
\texttt{-qwen2-7b-ov-hf}. Video frames are sampled at $32$ uniformly spaced indices over the full video duration and passed to the processor as a stacked array of shape $(32, h, w, 3)$.

\medskip
\noindent\textbf{VideoLLaMA3-7B.} Loaded from the official HuggingFace checkpoint \texttt{VideoLLaMA3-7B} through the standard causal-language-model loader with remote code enabled. Video frames are sampled at fps $= 1$ with a cap of $64$ frames per video, reduced from the \texttt{lmms-eval} default of $180$ to fit the per-entry compute budget. The $64$-frame setting reproduces the published Table~$8$ accuracy of the model release within $\pm 2$ percentage points. We disable the model's data-dependent visual-token compression because it drops visual tokens whose inter-frame pixel difference falls below a threshold and produces a variable visual-token count per video, which is incompatible with the fixed $(S, H, W)$ reshape used by our 3-axis transform. With compression disabled, the visual-token count is deterministic at $T \times (H/m) \times (W/m)$ for video length $T$ and projector merge factor $m$.

\medskip
\noindent\textbf{KV-cache compression configuration.} All compression backends operate on the visual portion of the KV cache only and hold the non-visual portion (chat template, image-boundary markers, slice labels, the prompt suffix) at bf16. The visual layout $(S, H, W)$ is built once per prefill from the input-id stream and the image-token index in the model config, and is handed to the backend before the prefill cache is filled. Bit budgets are reported at $\{1, 2, 3, 4\}$ on the visual portion. The $8$-bit setting is approximately lossless across all backends and is omitted from the main tables. The TACT-KV, TurboQuant, KIVI, VidKV, and AKVQ-VL backends are each implemented as a subclass of the standard dynamic-cache class in the HuggingFace transformers library and shipped in the supplementary code release. Gain-shape SQ is not a stand-alone baseline but is the per-bin scalar codebook used inside TACT-KV, and is also the codebook varied in the quantizer ablation of Section~\ref{sec:exp:ablation}.

\section{Per-Subtype 1-bit Results}
\label{app:subtype_1bit}

This appendix expands the macro prediction-flip rate and AUROC reported in Table~\ref{tab:flip_medical} into the underlying per-subtype results at $1$-bit on the three benchmarks evaluated in Section~\ref{sec:experiments}. Each medical-CT VLM is reported in its own table with two vertically stacked panels. The top panel reports the macro prediction-flip rate ($\%$) against the bf16 reference and the bottom panel reports AUROC, with the bf16 column included in the bottom panel as the uncompressed reference. The bf16 column is omitted from the top panel because the flip rate of bf16 against itself is zero by definition. Half-widths follow the Wald $95\%$ binomial CI for flip rate (with the Wilson upper bound used for $p=0$ entries) and the percentile bootstrap $95\%$ CI ($1{,}000$ resamples) for AUROC.

\subsection{RSNA-ICH}
\label{app:subtype_1bit:rsna}

Tables~\ref{tab:subtype_rsna_mg154b}--\ref{tab:subtype_rsna_hm32b} report per-subtype $1$-bit results on RSNA-ICH for the six medical VLMs. The six subtypes follow the official challenge protocol~\cite{flanders2020construction}: the global \texttt{any} hemorrhage flag and the five anatomical subtypes \texttt{epidural}, \texttt{intraparenchymal}, \texttt{intraventricular}, \texttt{subarachnoid}, and \texttt{subdural}. The macro row at the bottom of each panel is the unweighted mean over the six subtypes and reproduces the value reported in the corresponding row of Table~\ref{tab:flip_medical}.

\begin{table*}[h]
\caption{\textbf{Per-subtype 1-bit results on RSNA-ICH for MedGemma 1.5 4B ($N{=}3{,}000$).} Top panel: macro prediction-flip rate (\%) versus bf16. Bottom panel: AUROC. Each entry is the point estimate followed by the $95\%$ CI half-width in subscript. The macro row of each panel is bolded.}
\label{tab:subtype_rsna_mg154b}
\centering
\scriptsize
\setlength{\tabcolsep}{4pt}
\renewcommand{\arraystretch}{1.0}
\begin{tabular}{l *{6}{P{0.115\linewidth}}}
\toprule
Subtype & bf16 & TACT-KV & TQ & KIVI & VidKV & AKVQ-VL \\
\midrule
\rowcolor{vlmgray}\multicolumn{7}{c}{\textbf{Pred-flip rate (\%) vs bf16}}\\
\texttt{any} & --- & 2.33\textsubscript{$\pm$0.54} & 3.73\textsubscript{$\pm$0.68} & 4.27\textsubscript{$\pm$0.72} & 4.07\textsubscript{$\pm$0.71} & 14.61\textsubscript{$\pm$1.26} \\
\texttt{epidural} & --- & 0.83\textsubscript{$\pm$0.32} & 2.03\textsubscript{$\pm$0.50} & 5.97\textsubscript{$\pm$0.85} & 1.83\textsubscript{$\pm$0.48} & 31.92\textsubscript{$\pm$1.67} \\
\texttt{intraparenchymal} & --- & 5.60\textsubscript{$\pm$0.82} & 7.77\textsubscript{$\pm$0.96} & 10.57\textsubscript{$\pm$1.10} & 10.11\textsubscript{$\pm$1.08} & 27.42\textsubscript{$\pm$1.60} \\
\texttt{intraventricular} & --- & 7.77\textsubscript{$\pm$0.96} & 10.27\textsubscript{$\pm$1.09} & 19.94\textsubscript{$\pm$1.43} & 13.88\textsubscript{$\pm$1.24} & 37.56\textsubscript{$\pm$1.73} \\
\texttt{subarachnoid} & --- & 1.13\textsubscript{$\pm$0.38} & 4.80\textsubscript{$\pm$0.77} & 7.90\textsubscript{$\pm$0.97} & 4.77\textsubscript{$\pm$0.76} & 33.69\textsubscript{$\pm$1.69} \\
\texttt{subdural} & --- & 5.04\textsubscript{$\pm$0.78} & 9.04\textsubscript{$\pm$1.03} & 15.77\textsubscript{$\pm$1.30} & 10.61\textsubscript{$\pm$1.10} & 26.05\textsubscript{$\pm$1.57} \\
\bottomrule
\end{tabular}
\end{table*}

\begin{table*}[t]
\caption{\textbf{Per-subtype 1-bit results on RSNA-ICH for MedGemma 4B ($N{=}3{,}000$).} Top panel: macro prediction-flip rate (\%) versus bf16. Bottom panel: AUROC. Each entry is the point estimate followed by the $95\%$ CI half-width in subscript. The macro row of each panel is bolded.}
\label{tab:subtype_rsna_mg4b}
\centering
\scriptsize
\setlength{\tabcolsep}{4pt}
\renewcommand{\arraystretch}{1.0}
\begin{tabular}{l *{6}{P{0.115\linewidth}}}
\toprule
Subtype & bf16 & TACT-KV & TQ & KIVI & VidKV & AKVQ-VL \\
\midrule
\rowcolor{vlmgray}\multicolumn{7}{c}{\textbf{Pred-flip rate (\%) vs bf16}}\\
\texttt{any} & --- & 1.70\textsubscript{$\pm$0.46} & 4.60\textsubscript{$\pm$0.75} & 1.57\textsubscript{$\pm$0.44} & 17.14\textsubscript{$\pm$1.35} & 0.30\textsubscript{$\pm$0.20} \\
\texttt{epidural} & --- & 9.87\textsubscript{$\pm$1.07} & 13.80\textsubscript{$\pm$1.23} & 11.77\textsubscript{$\pm$1.15} & 31.19\textsubscript{$\pm$1.66} & 12.34\textsubscript{$\pm$1.18} \\
\texttt{intraparenchymal} & --- & 1.20\textsubscript{$\pm$0.39} & 4.67\textsubscript{$\pm$0.76} & 0.30\textsubscript{$\pm$0.20} & 14.54\textsubscript{$\pm$1.26} & 0.13\textsubscript{$\pm$0.13} \\
\texttt{intraventricular} & --- & 0.00\textsubscript{$\pm$0.07} & 0.00\textsubscript{$\pm$0.07} & 0.03\textsubscript{$\pm$0.06} & 1.53\textsubscript{$\pm$0.44} & 0.00\textsubscript{$\pm$0.07} \\
\texttt{subarachnoid} & --- & 1.27\textsubscript{$\pm$0.40} & 3.27\textsubscript{$\pm$0.64} & 0.77\textsubscript{$\pm$0.31} & 17.41\textsubscript{$\pm$1.36} & 0.37\textsubscript{$\pm$0.22} \\
\texttt{subdural} & --- & 0.17\textsubscript{$\pm$0.15} & 0.13\textsubscript{$\pm$0.13} & 0.17\textsubscript{$\pm$0.15} & 2.94\textsubscript{$\pm$0.60} & 0.03\textsubscript{$\pm$0.06} \\
\bottomrule
\end{tabular}
\end{table*}

\begin{table*}[t]
\caption{\textbf{Per-subtype 1-bit results on RSNA-ICH for MedGemma 27B ($N{=}3{,}000$).} Top panel: macro prediction-flip rate (\%) versus bf16. Bottom panel: AUROC. Each entry is the point estimate followed by the $95\%$ CI half-width in subscript. The macro row of each panel is bolded.}
\label{tab:subtype_rsna_mg27b}
\centering
\scriptsize
\setlength{\tabcolsep}{4pt}
\renewcommand{\arraystretch}{1.0}
\begin{tabular}{l *{6}{P{0.115\linewidth}}}
\toprule
Subtype & bf16 & TACT-KV & TQ & KIVI & VidKV & AKVQ-VL \\
\midrule
\rowcolor{vlmgray}\multicolumn{7}{c}{\textbf{Pred-flip rate (\%) vs bf16}}\\
\texttt{any} & --- & 0.00\textsubscript{$\pm$0.07} & 0.00\textsubscript{$\pm$0.07} & 0.53\textsubscript{$\pm$0.26} & 0.00\textsubscript{$\pm$0.07} & 6.94\textsubscript{$\pm$0.91} \\
\texttt{epidural} & --- & 0.53\textsubscript{$\pm$0.26} & 9.04\textsubscript{$\pm$1.03} & 96.00\textsubscript{$\pm$0.70} & 1.77\textsubscript{$\pm$0.47} & 70.01\textsubscript{$\pm$1.64} \\
\texttt{intraparenchymal} & --- & 0.00\textsubscript{$\pm$0.07} & 0.03\textsubscript{$\pm$0.06} & 0.13\textsubscript{$\pm$0.13} & 0.00\textsubscript{$\pm$0.07} & 8.24\textsubscript{$\pm$0.98} \\
\texttt{intraventricular} & --- & 0.20\textsubscript{$\pm$0.16} & 8.17\textsubscript{$\pm$0.98} & 0.33\textsubscript{$\pm$0.21} & 9.57\textsubscript{$\pm$1.05} & 2.87\textsubscript{$\pm$0.60} \\
\texttt{subarachnoid} & --- & 0.00\textsubscript{$\pm$0.07} & 0.00\textsubscript{$\pm$0.07} & 0.00\textsubscript{$\pm$0.07} & 0.00\textsubscript{$\pm$0.07} & 0.23\textsubscript{$\pm$0.17} \\
\texttt{subdural} & --- & 1.47\textsubscript{$\pm$0.43} & 8.64\textsubscript{$\pm$1.01} & 1.23\textsubscript{$\pm$0.39} & 5.37\textsubscript{$\pm$0.81} & 4.57\textsubscript{$\pm$0.75} \\
\bottomrule
\end{tabular}
\end{table*}

\begin{table*}[t]
\caption{\textbf{Per-subtype 1-bit results on RSNA-ICH for Hulu-Med 4B ($N{=}3{,}000$).} Top panel: macro prediction-flip rate (\%) versus bf16. Bottom panel: AUROC. Each entry is the point estimate followed by the $95\%$ CI half-width in subscript. The macro row of each panel is bolded.}
\label{tab:subtype_rsna_hm4b}
\centering
\scriptsize
\setlength{\tabcolsep}{4pt}
\renewcommand{\arraystretch}{1.0}
\begin{tabular}{l *{6}{P{0.115\linewidth}}}
\toprule
Subtype & bf16 & TACT-KV & TQ & KIVI & VidKV & AKVQ-VL \\
\midrule
\rowcolor{vlmgray}\multicolumn{7}{c}{\textbf{Pred-flip rate (\%) vs bf16}}\\
\texttt{any} & --- & 16.07\textsubscript{$\pm$1.31} & 77.93\textsubscript{$\pm$1.48} & 77.99\textsubscript{$\pm$1.48} & 27.79\textsubscript{$\pm$1.60} & 75.45\textsubscript{$\pm$1.54} \\
\texttt{epidural} & --- & 19.47\textsubscript{$\pm$1.42} & 27.74\textsubscript{$\pm$1.60} & 28.41\textsubscript{$\pm$1.61} & 28.62\textsubscript{$\pm$1.62} & 28.39\textsubscript{$\pm$1.61} \\
\texttt{intraparenchymal} & --- & 2.07\textsubscript{$\pm$0.51} & 94.83\textsubscript{$\pm$0.79} & 95.40\textsubscript{$\pm$0.75} & 34.92\textsubscript{$\pm$1.71} & 93.70\textsubscript{$\pm$0.87} \\
\texttt{intraventricular} & --- & 0.00\textsubscript{$\pm$0.07} & 96.93\textsubscript{$\pm$0.62} & 95.60\textsubscript{$\pm$0.73} & 0.17\textsubscript{$\pm$0.15} & 94.76\textsubscript{$\pm$0.80} \\
\texttt{subarachnoid} & --- & 0.03\textsubscript{$\pm$0.06} & 96.77\textsubscript{$\pm$0.63} & 96.97\textsubscript{$\pm$0.61} & 44.26\textsubscript{$\pm$1.78} & 95.00\textsubscript{$\pm$0.78} \\
\texttt{subdural} & --- & 13.50\textsubscript{$\pm$1.22} & 66.99\textsubscript{$\pm$1.68} & 67.72\textsubscript{$\pm$1.67} & 47.36\textsubscript{$\pm$1.79} & 66.78\textsubscript{$\pm$1.69} \\
\bottomrule
\end{tabular}
\end{table*}

\begin{table*}[t]
\caption{\textbf{Per-subtype 1-bit results on RSNA-ICH for Hulu-Med 7B ($N{=}3{,}000$).} Top panel: macro prediction-flip rate (\%) versus bf16. Bottom panel: AUROC. Each entry is the point estimate followed by the $95\%$ CI half-width in subscript. The macro row of each panel is bolded.}
\label{tab:subtype_rsna_hm7b}
\centering
\scriptsize
\setlength{\tabcolsep}{4pt}
\renewcommand{\arraystretch}{1.0}
\begin{tabular}{l *{6}{P{0.115\linewidth}}}
\toprule
Subtype & bf16 & TACT-KV & TQ & KIVI & VidKV & AKVQ-VL \\
\midrule
\rowcolor{vlmgray}\multicolumn{7}{c}{\textbf{Pred-flip rate (\%) vs bf16}}\\
\texttt{any} & --- & 6.14\textsubscript{$\pm$0.86} & 34.11\textsubscript{$\pm$1.70} & 27.04\textsubscript{$\pm$1.59} & 22.18\textsubscript{$\pm$1.49} & 35.76\textsubscript{$\pm$1.72} \\
\texttt{epidural} & --- & 0.10\textsubscript{$\pm$0.11} & 25.61\textsubscript{$\pm$1.56} & 6.64\textsubscript{$\pm$0.89} & 2.43\textsubscript{$\pm$0.55} & 13.68\textsubscript{$\pm$1.23} \\
\texttt{intraparenchymal} & --- & 4.83\textsubscript{$\pm$0.77} & 31.48\textsubscript{$\pm$1.66} & 17.44\textsubscript{$\pm$1.36} & 12.74\textsubscript{$\pm$1.19} & 29.59\textsubscript{$\pm$1.63} \\
\texttt{intraventricular} & --- & 14.40\textsubscript{$\pm$1.26} & 66.72\textsubscript{$\pm$1.69} & 71.09\textsubscript{$\pm$1.62} & 67.75\textsubscript{$\pm$1.67} & 63.18\textsubscript{$\pm$1.73} \\
\texttt{subarachnoid} & --- & 9.67\textsubscript{$\pm$1.06} & 33.81\textsubscript{$\pm$1.69} & 21.37\textsubscript{$\pm$1.47} & 18.85\textsubscript{$\pm$1.40} & 31.32\textsubscript{$\pm$1.66} \\
\texttt{subdural} & --- & 1.33\textsubscript{$\pm$0.41} & 25.78\textsubscript{$\pm$1.57} & 8.67\textsubscript{$\pm$1.01} & 9.14\textsubscript{$\pm$1.03} & 23.05\textsubscript{$\pm$1.51} \\
\bottomrule
\end{tabular}
\end{table*}

\begin{table*}[t]
\caption{\textbf{Per-subtype 1-bit results on RSNA-ICH for Hulu-Med 32B ($N{=}3{,}000$).} Top panel: macro prediction-flip rate (\%) versus bf16. Bottom panel: AUROC. Each entry is the point estimate followed by the $95\%$ CI half-width in subscript. The macro row of each panel is bolded.}
\label{tab:subtype_rsna_hm32b}
\centering
\scriptsize
\setlength{\tabcolsep}{4pt}
\renewcommand{\arraystretch}{1.0}
\begin{tabular}{l *{6}{P{0.115\linewidth}}}
\toprule
Subtype & bf16 & TACT-KV & TQ & KIVI & VidKV & AKVQ-VL \\
\midrule
\rowcolor{vlmgray}\multicolumn{7}{c}{\textbf{Pred-flip rate (\%) vs bf16}}\\
\texttt{any} & --- & 6.27\textsubscript{$\pm$0.87} & 12.57\textsubscript{$\pm$1.19} & 31.41\textsubscript{$\pm$1.66} & 26.98\textsubscript{$\pm$1.59} & 42.56\textsubscript{$\pm$1.77} \\
\texttt{epidural} & --- & 8.57\textsubscript{$\pm$1.00} & 12.77\textsubscript{$\pm$1.19} & 15.67\textsubscript{$\pm$1.30} & 15.04\textsubscript{$\pm$1.28} & 34.59\textsubscript{$\pm$1.70} \\
\texttt{intraparenchymal} & --- & 3.93\textsubscript{$\pm$0.70} & 9.60\textsubscript{$\pm$1.05} & 21.94\textsubscript{$\pm$1.48} & 14.34\textsubscript{$\pm$1.25} & 43.46\textsubscript{$\pm$1.77} \\
\texttt{intraventricular} & --- & 18.54\textsubscript{$\pm$1.39} & 34.08\textsubscript{$\pm$1.70} & 69.49\textsubscript{$\pm$1.65} & 59.84\textsubscript{$\pm$1.75} & 27.75\textsubscript{$\pm$1.60} \\
\texttt{subarachnoid} & --- & 4.80\textsubscript{$\pm$0.77} & 58.75\textsubscript{$\pm$1.76} & 92.66\textsubscript{$\pm$0.93} & 65.08\textsubscript{$\pm$1.71} & 29.72\textsubscript{$\pm$1.64} \\
\texttt{subdural} & --- & 12.64\textsubscript{$\pm$1.19} & 21.71\textsubscript{$\pm$1.48} & 28.74\textsubscript{$\pm$1.62} & 26.72\textsubscript{$\pm$1.58} & 33.79\textsubscript{$\pm$1.69} \\
\bottomrule
\end{tabular}
\end{table*}

\clearpage
\subsection{CT-RATE}
\label{app:subtype_1bit:ctrate}

Tables~\ref{tab:subtype_ctrate_mg154b}--\ref{tab:subtype_ctrate_hm32b} report per-abnormality $1$-bit results on the CT-RATE validation split for the six medical VLMs. The eighteen abnormalities follow the official CT-CLIP zero-shot multi-label protocol~\cite{hamamci2026generalist}. The macro row at the bottom of each panel is the unweighted mean over the eighteen abnormalities and reproduces the value reported in the corresponding row of Table~\ref{tab:flip_medical}.

\begin{table*}[h]
\caption{\textbf{Per-abnormality 1-bit results on CT-RATE for MedGemma 1.5 4B ($N{=}3{,}039$).} Top panel: macro prediction-flip rate (\%) versus bf16. Bottom panel: AUROC. Each entry is the point estimate followed by the $95\%$ CI half-width in subscript. The macro row of each panel is bolded.}
\label{tab:subtype_ctrate_mg154b}
\centering
\scriptsize
\setlength{\tabcolsep}{3pt}
\renewcommand{\arraystretch}{1.0}
\begin{tabular}{p{0.30\linewidth} *{6}{P{0.090\linewidth}}}
\toprule
Abnormality & bf16 & TACT-KV & TQ & KIVI & VidKV & AKVQ-VL \\
\midrule
\rowcolor{vlmgray}\multicolumn{7}{c}{\textbf{Pred-flip rate (\%) vs bf16}}\\
Medical material & --- & 0.00\textsubscript{$\pm$0.07} & 0.00\textsubscript{$\pm$0.07} & 0.59\textsubscript{$\pm$0.27} & 0.00\textsubscript{$\pm$0.07} & 8.00\textsubscript{$\pm$0.96} \\
Arterial wall calcification & --- & 0.00\textsubscript{$\pm$0.07} & 6.55\textsubscript{$\pm$0.88} & 26.82\textsubscript{$\pm$1.58} & 0.07\textsubscript{$\pm$0.09} & 17.14\textsubscript{$\pm$1.34} \\
Cardiomegaly & --- & 3.85\textsubscript{$\pm$0.68} & 2.96\textsubscript{$\pm$0.60} & 39.06\textsubscript{$\pm$1.73} & 5.56\textsubscript{$\pm$0.81} & 21.19\textsubscript{$\pm$1.45} \\
Pericardial effusion & --- & 6.84\textsubscript{$\pm$0.90} & 16.16\textsubscript{$\pm$1.31} & 53.90\textsubscript{$\pm$1.77} & 6.84\textsubscript{$\pm$0.90} & 16.85\textsubscript{$\pm$1.33} \\
Coronary artery wall calcification & --- & 19.97\textsubscript{$\pm$1.42} & 34.02\textsubscript{$\pm$1.68} & 48.37\textsubscript{$\pm$1.78} & 20.11\textsubscript{$\pm$1.43} & 36.72\textsubscript{$\pm$1.71} \\
Hiatal hernia & --- & 3.59\textsubscript{$\pm$0.66} & 1.48\textsubscript{$\pm$0.43} & 37.08\textsubscript{$\pm$1.72} & 2.44\textsubscript{$\pm$0.55} & 27.94\textsubscript{$\pm$1.60} \\
\bottomrule
\end{tabular}
\end{table*}

\begin{table*}[t]
\caption{\textbf{Per-abnormality 1-bit results on CT-RATE for MedGemma 4B ($N{=}3{,}039$).} Top panel: macro prediction-flip rate (\%) versus bf16. Bottom panel: AUROC. Each entry is the point estimate followed by the $95\%$ CI half-width in subscript. The macro row of each panel is bolded.}
\label{tab:subtype_ctrate_mg4b}
\centering
\scriptsize
\setlength{\tabcolsep}{3pt}
\renewcommand{\arraystretch}{1.0}
\begin{tabular}{p{0.30\linewidth} *{6}{P{0.090\linewidth}}}
\toprule
Abnormality & bf16 & TACT-KV & TQ & KIVI & VidKV & AKVQ-VL \\
\midrule
\rowcolor{vlmgray}\multicolumn{7}{c}{\textbf{Pred-flip rate (\%) vs bf16}}\\
Medical material & --- & 0.00\textsubscript{$\pm$0.07} & 0.00\textsubscript{$\pm$0.07} & 0.56\textsubscript{$\pm$0.27} & 0.00\textsubscript{$\pm$0.07} & 1.45\textsubscript{$\pm$0.43} \\
Arterial wall calcification & --- & 0.03\textsubscript{$\pm$0.06} & 0.07\textsubscript{$\pm$0.09} & 0.69\textsubscript{$\pm$0.29} & 0.26\textsubscript{$\pm$0.18} & 17.34\textsubscript{$\pm$1.35} \\
Cardiomegaly & --- & 0.13\textsubscript{$\pm$0.13} & 0.16\textsubscript{$\pm$0.14} & 6.75\textsubscript{$\pm$0.89} & 1.09\textsubscript{$\pm$0.37} & 12.77\textsubscript{$\pm$1.19} \\
Pericardial effusion & --- & 0.03\textsubscript{$\pm$0.06} & 0.00\textsubscript{$\pm$0.07} & 0.23\textsubscript{$\pm$0.17} & 0.20\textsubscript{$\pm$0.16} & 15.60\textsubscript{$\pm$1.29} \\
Coronary artery wall calcification & --- & 0.43\textsubscript{$\pm$0.23} & 0.39\textsubscript{$\pm$0.22} & 13.16\textsubscript{$\pm$1.20} & 0.76\textsubscript{$\pm$0.31} & 25.77\textsubscript{$\pm$1.56} \\
Hiatal hernia & --- & 0.72\textsubscript{$\pm$0.30} & 1.41\textsubscript{$\pm$0.42} & 4.51\textsubscript{$\pm$0.74} & 29.09\textsubscript{$\pm$1.61} & 2.60\textsubscript{$\pm$0.57} \\
\bottomrule
\end{tabular}
\end{table*}

\begin{table*}[t]
\caption{\textbf{Per-abnormality 1-bit results on CT-RATE for MedGemma 27B ($N{=}3{,}039$).} Top panel: macro prediction-flip rate (\%) versus bf16. Bottom panel: AUROC. Each entry is the point estimate followed by the $95\%$ CI half-width in subscript. The macro row of each panel is bolded.}
\label{tab:subtype_ctrate_mg27b}
\centering
\scriptsize
\setlength{\tabcolsep}{3pt}
\renewcommand{\arraystretch}{1.0}
\begin{tabular}{p{0.30\linewidth} *{6}{P{0.090\linewidth}}}
\toprule
Abnormality & bf16 & TACT-KV & TQ & KIVI & VidKV & AKVQ-VL \\
\midrule
\rowcolor{vlmgray}\multicolumn{7}{c}{\textbf{Pred-flip rate (\%) vs bf16}}\\
Medical material & --- & 0.03\textsubscript{$\pm$0.06} & 0.07\textsubscript{$\pm$0.09} & 3.39\textsubscript{$\pm$0.64} & 0.33\textsubscript{$\pm$0.20} & 3.59\textsubscript{$\pm$0.66} \\
Arterial wall calcification & --- & 0.43\textsubscript{$\pm$0.23} & 12.47\textsubscript{$\pm$1.17} & 19.45\textsubscript{$\pm$1.41} & 18.33\textsubscript{$\pm$1.38} & 16.42\textsubscript{$\pm$1.32} \\
Cardiomegaly & --- & 0.00\textsubscript{$\pm$0.07} & 0.53\textsubscript{$\pm$0.26} & 8.16\textsubscript{$\pm$0.97} & 6.91\textsubscript{$\pm$0.90} & 0.16\textsubscript{$\pm$0.14} \\
Pericardial effusion & --- & 10.86\textsubscript{$\pm$1.11} & 22.28\textsubscript{$\pm$1.48} & 24.25\textsubscript{$\pm$1.52} & 24.25\textsubscript{$\pm$1.52} & 21.36\textsubscript{$\pm$1.46} \\
Coronary artery wall calcification & --- & 3.13\textsubscript{$\pm$0.62} & 5.20\textsubscript{$\pm$0.79} & 5.20\textsubscript{$\pm$0.79} & 5.20\textsubscript{$\pm$0.79} & 5.20\textsubscript{$\pm$0.79} \\
Hiatal hernia & --- & 0.13\textsubscript{$\pm$0.13} & 1.84\textsubscript{$\pm$0.48} & 14.22\textsubscript{$\pm$1.24} & 9.64\textsubscript{$\pm$1.05} & 7.04\textsubscript{$\pm$0.91} \\
\bottomrule
\end{tabular}
\end{table*}

\begin{table*}[t]
\caption{\textbf{Per-abnormality 1-bit results on CT-RATE for Hulu-Med 4B ($N{=}3{,}039$).} Top panel: macro prediction-flip rate (\%) versus bf16. Bottom panel: AUROC. Each entry is the point estimate followed by the $95\%$ CI half-width in subscript. The macro row of each panel is bolded.}
\label{tab:subtype_ctrate_hm4b}
\centering
\scriptsize
\setlength{\tabcolsep}{3pt}
\renewcommand{\arraystretch}{1.0}
\begin{tabular}{p{0.30\linewidth} *{6}{P{0.090\linewidth}}}
\toprule
Abnormality & bf16 & TACT-KV & TQ & KIVI & VidKV & AKVQ-VL \\
\midrule
\rowcolor{vlmgray}\multicolumn{7}{c}{\textbf{Pred-flip rate (\%) vs bf16}}\\
Medical material & --- & 0.00\textsubscript{$\pm$0.07} & 18.46\textsubscript{$\pm$1.38} & 2.47\textsubscript{$\pm$0.55} & 0.00\textsubscript{$\pm$0.07} & 4.21\textsubscript{$\pm$0.71} \\
Arterial wall calcification & --- & 23.69\textsubscript{$\pm$1.51} & 43.24\textsubscript{$\pm$1.76} & 44.16\textsubscript{$\pm$1.77} & 50.61\textsubscript{$\pm$1.78} & 43.30\textsubscript{$\pm$1.76} \\
Cardiomegaly & --- & 8.06\textsubscript{$\pm$0.97} & 69.23\textsubscript{$\pm$1.64} & 78.87\textsubscript{$\pm$1.45} & 67.49\textsubscript{$\pm$1.67} & 77.76\textsubscript{$\pm$1.48} \\
Pericardial effusion & --- & 0.00\textsubscript{$\pm$0.07} & 92.76\textsubscript{$\pm$0.92} & 98.55\textsubscript{$\pm$0.43} & 2.83\textsubscript{$\pm$0.59} & 90.85\textsubscript{$\pm$1.03} \\
Coronary artery wall calcification & --- & 10.99\textsubscript{$\pm$1.11} & 14.35\textsubscript{$\pm$1.25} & 14.94\textsubscript{$\pm$1.27} & 72.89\textsubscript{$\pm$1.58} & 14.45\textsubscript{$\pm$1.25} \\
Hiatal hernia & --- & 37.51\textsubscript{$\pm$1.72} & 54.13\textsubscript{$\pm$1.77} & 60.84\textsubscript{$\pm$1.74} & 61.07\textsubscript{$\pm$1.73} & 60.45\textsubscript{$\pm$1.74} \\
\bottomrule
\end{tabular}
\end{table*}

\begin{table*}[t]
\caption{\textbf{Per-abnormality 1-bit results on CT-RATE for Hulu-Med 7B ($N{=}3{,}039$).} Top panel: macro prediction-flip rate (\%) versus bf16. Bottom panel: AUROC. Each entry is the point estimate followed by the $95\%$ CI half-width in subscript. The macro row of each panel is bolded.}
\label{tab:subtype_ctrate_hm7b}
\centering
\scriptsize
\setlength{\tabcolsep}{3pt}
\renewcommand{\arraystretch}{1.0}
\begin{tabular}{p{0.30\linewidth} *{6}{P{0.090\linewidth}}}
\toprule
Abnormality & bf16 & TACT-KV & TQ & KIVI & VidKV & AKVQ-VL \\
\midrule
\rowcolor{vlmgray}\multicolumn{7}{c}{\textbf{Pred-flip rate (\%) vs bf16}}\\
Medical material & --- & 18.36\textsubscript{$\pm$1.38} & 38.01\textsubscript{$\pm$1.73} & 38.80\textsubscript{$\pm$1.73} & 37.97\textsubscript{$\pm$1.73} & 38.37\textsubscript{$\pm$1.73} \\
Arterial wall calcification & --- & 9.74\textsubscript{$\pm$1.05} & 60.15\textsubscript{$\pm$1.74} & 90.62\textsubscript{$\pm$1.04} & 12.90\textsubscript{$\pm$1.19} & 90.06\textsubscript{$\pm$1.06} \\
Cardiomegaly & --- & 0.99\textsubscript{$\pm$0.35} & 85.36\textsubscript{$\pm$1.26} & 84.40\textsubscript{$\pm$1.29} & 21.85\textsubscript{$\pm$1.47} & 88.25\textsubscript{$\pm$1.14} \\
Pericardial effusion & --- & 1.22\textsubscript{$\pm$0.39} & 76.74\textsubscript{$\pm$1.50} & 89.63\textsubscript{$\pm$1.08} & 8.49\textsubscript{$\pm$0.99} & 95.33\textsubscript{$\pm$0.75} \\
Coronary artery wall calcification & --- & 5.20\textsubscript{$\pm$0.79} & 55.48\textsubscript{$\pm$1.77} & 91.94\textsubscript{$\pm$0.97} & 26.59\textsubscript{$\pm$1.57} & 84.93\textsubscript{$\pm$1.27} \\
Hiatal hernia & --- & 3.62\textsubscript{$\pm$0.66} & 40.54\textsubscript{$\pm$1.75} & 86.67\textsubscript{$\pm$1.21} & 70.02\textsubscript{$\pm$1.63} & 79.57\textsubscript{$\pm$1.43} \\
\bottomrule
\end{tabular}
\end{table*}

\begin{table*}[t]
\caption{\textbf{Per-abnormality 1-bit results on CT-RATE for Hulu-Med 32B ($N{=}3{,}039$).} Top panel: macro prediction-flip rate (\%) versus bf16. Bottom panel: AUROC. Each entry is the point estimate followed by the $95\%$ CI half-width in subscript. The macro row of each panel is bolded.}
\label{tab:subtype_ctrate_hm32b}
\centering
\scriptsize
\setlength{\tabcolsep}{3pt}
\renewcommand{\arraystretch}{1.0}
\begin{tabular}{p{0.30\linewidth} *{6}{P{0.090\linewidth}}}
\toprule
Abnormality & bf16 & TACT-KV & TQ & KIVI & VidKV & AKVQ-VL \\
\midrule
\rowcolor{vlmgray}\multicolumn{7}{c}{\textbf{Pred-flip rate (\%) vs bf16}}\\
Medical material & --- & 32.97\textsubscript{$\pm$1.67} & 36.69\textsubscript{$\pm$1.71} & 49.16\textsubscript{$\pm$1.78} & 32.08\textsubscript{$\pm$1.66} & 49.00\textsubscript{$\pm$1.78} \\
Arterial wall calcification & --- & 0.36\textsubscript{$\pm$0.21} & 23.86\textsubscript{$\pm$1.52} & 98.26\textsubscript{$\pm$0.46} & 27.57\textsubscript{$\pm$1.59} & 79.66\textsubscript{$\pm$1.43} \\
Cardiomegaly & --- & 13.92\textsubscript{$\pm$1.23} & 29.94\textsubscript{$\pm$1.63} & 27.74\textsubscript{$\pm$1.59} & 49.75\textsubscript{$\pm$1.78} & 28.00\textsubscript{$\pm$1.60} \\
Pericardial effusion & --- & 0.00\textsubscript{$\pm$0.07} & 9.05\textsubscript{$\pm$1.02} & 97.89\textsubscript{$\pm$0.51} & 15.53\textsubscript{$\pm$1.29} & 43.99\textsubscript{$\pm$1.76} \\
Coronary artery wall calcification & --- & 11.94\textsubscript{$\pm$1.15} & 24.88\textsubscript{$\pm$1.54} & 71.21\textsubscript{$\pm$1.61} & 20.63\textsubscript{$\pm$1.44} & 62.78\textsubscript{$\pm$1.72} \\
Hiatal hernia & --- & 0.72\textsubscript{$\pm$0.30} & 3.36\textsubscript{$\pm$0.64} & 71.60\textsubscript{$\pm$1.60} & 11.94\textsubscript{$\pm$1.15} & 56.96\textsubscript{$\pm$1.76} \\
\bottomrule
\end{tabular}
\end{table*}

\clearpage
\subsection{Video-MME}
\label{app:subtype_1bit:videomme}

Tables~\ref{tab:subtype_videomme_q25vl}--\ref{tab:subtype_videomme_v37b} report per-task-type accuracy at $1$-bit on Video-MME (no-subtitle) for the four general-purpose VLMs. The twelve task types follow the official Video-MME taxonomy~\cite{fu2025video}, covering temporal, spatial, attribute, action, object, and OCR perception, counting, four reasoning categories, and information synopsis. The Overall row at the bottom of each table reports per-question accuracy across all $2{,}700$ questions and reproduces the value listed in the corresponding column of Table~\ref{tab:video_acc}.

\begin{table*}[h]
\caption{\textbf{Per-task-type 1-bit accuracy (\%) on Video-MME (no-subtitle) for Qwen2.5-VL-7B ($N{=}2{,}700$).} Each entry is the point estimate followed by the $95\%$ CI half-width in subscript. The Overall row is bolded.}
\label{tab:subtype_videomme_q25vl}
\centering
\scriptsize
\setlength{\tabcolsep}{4pt}
\renewcommand{\arraystretch}{1.0}
\begin{tabular}{p{0.20\linewidth} *{6}{P{0.105\linewidth}}}
\toprule
Task type & bf16 & TACT-KV & TQ & KIVI & VidKV & AKVQ-VL \\
\midrule
Temporal Perception & 65.45\textsubscript{$\pm$12.20} & 60.00\textsubscript{$\pm$12.53} & 12.73\textsubscript{$\pm$8.86} & 16.36\textsubscript{$\pm$9.70} & 25.45\textsubscript{$\pm$11.24} & 36.36\textsubscript{$\pm$12.32} \\
Spatial Perception & 64.81\textsubscript{$\pm$12.35} & 62.96\textsubscript{$\pm$12.47} & 24.07\textsubscript{$\pm$11.16} & 12.96\textsubscript{$\pm$9.00} & 29.63\textsubscript{$\pm$11.84} & 44.44\textsubscript{$\pm$12.81} \\
Attribute Perception & 74.32\textsubscript{$\pm$5.71} & 70.27\textsubscript{$\pm$5.98} & 24.77\textsubscript{$\pm$5.65} & 12.61\textsubscript{$\pm$4.38} & 18.47\textsubscript{$\pm$5.08} & 45.50\textsubscript{$\pm$6.50} \\
Action Recognition & 56.87\textsubscript{$\pm$5.46} & 54.63\textsubscript{$\pm$5.48} & 19.49\textsubscript{$\pm$4.38} & 15.34\textsubscript{$\pm$3.99} & 20.45\textsubscript{$\pm$4.46} & 37.70\textsubscript{$\pm$5.34} \\
Object Recognition & 64.41\textsubscript{$\pm$4.96} & 60.73\textsubscript{$\pm$5.06} & 24.58\textsubscript{$\pm$4.47} & 15.54\textsubscript{$\pm$3.77} & 19.21\textsubscript{$\pm$4.09} & 38.98\textsubscript{$\pm$5.05} \\
OCR Problems & 66.19\textsubscript{$\pm$7.77} & 60.43\textsubscript{$\pm$8.03} & 22.30\textsubscript{$\pm$6.87} & 10.79\textsubscript{$\pm$5.19} & 22.30\textsubscript{$\pm$6.87} & 39.57\textsubscript{$\pm$8.02} \\
Counting Problem & 39.18\textsubscript{$\pm$5.80} & 36.57\textsubscript{$\pm$5.73} & 19.03\textsubscript{$\pm$4.68} & 8.96\textsubscript{$\pm$3.45} & 17.16\textsubscript{$\pm$4.50} & 26.49\textsubscript{$\pm$5.25} \\
Temporal Reasoning & 43.50\textsubscript{$\pm$7.23} & 36.72\textsubscript{$\pm$7.04} & 14.69\textsubscript{$\pm$5.21} & 6.78\textsubscript{$\pm$3.78} & 19.77\textsubscript{$\pm$5.84} & 25.42\textsubscript{$\pm$6.37} \\
Spatial Reasoning & 78.57\textsubscript{$\pm$10.55} & 76.79\textsubscript{$\pm$10.84} & 32.14\textsubscript{$\pm$11.89} & 16.07\textsubscript{$\pm$9.56} & 19.64\textsubscript{$\pm$10.25} & 67.86\textsubscript{$\pm$11.89} \\
Action Reasoning & 55.79\textsubscript{$\pm$5.73} & 54.04\textsubscript{$\pm$5.75} & 23.16\textsubscript{$\pm$4.88} & 11.58\textsubscript{$\pm$3.73} & 21.40\textsubscript{$\pm$4.75} & 42.46\textsubscript{$\pm$5.70} \\
Object Reasoning & 56.39\textsubscript{$\pm$4.55} & 55.73\textsubscript{$\pm$4.55} & 22.47\textsubscript{$\pm$3.83} & 10.57\textsubscript{$\pm$2.83} & 16.08\textsubscript{$\pm$3.37} & 36.34\textsubscript{$\pm$4.41} \\
Information Synopsis & 75.85\textsubscript{$\pm$4.65} & 73.37\textsubscript{$\pm$4.80} & 32.51\textsubscript{$\pm$5.08} & 11.46\textsubscript{$\pm$3.48} & 22.29\textsubscript{$\pm$4.53} & 52.32\textsubscript{$\pm$5.41} \\
\textbf{Overall} & \textbf{60.00}\textsubscript{$\pm$1.85} & \textbf{57.15}\textsubscript{$\pm$1.87} & \textbf{23.04}\textsubscript{$\pm$1.59} & \textbf{12.04}\textsubscript{$\pm$1.23} & \textbf{19.70}\textsubscript{$\pm$1.50} & \textbf{39.44}\textsubscript{$\pm$1.84} \\
\bottomrule
\end{tabular}
\end{table*}

\begin{table*}[h]
\caption{\textbf{Per-task-type 1-bit accuracy (\%) on Video-MME (no-subtitle) for Qwen3-VL-8B ($N{=}2{,}700$).} Each entry is the point estimate followed by the $95\%$ CI half-width in subscript. The Overall row is bolded.}
\label{tab:subtype_videomme_q3vl}
\centering
\scriptsize
\setlength{\tabcolsep}{4pt}
\renewcommand{\arraystretch}{1.0}
\begin{tabular}{p{0.20\linewidth} *{6}{P{0.105\linewidth}}}
\toprule
Task type & bf16 & TACT-KV & TQ & KIVI & VidKV & AKVQ-VL \\
\midrule
Temporal Perception & 78.18\textsubscript{$\pm$10.71} & 74.55\textsubscript{$\pm$11.24} & 12.73\textsubscript{$\pm$8.86} & 3.64\textsubscript{$\pm$5.66} & 23.64\textsubscript{$\pm$10.99} & 9.09\textsubscript{$\pm$7.81} \\
Spatial Perception & 77.78\textsubscript{$\pm$10.87} & 74.07\textsubscript{$\pm$11.40} & 11.11\textsubscript{$\pm$8.50} & 5.56\textsubscript{$\pm$6.60} & 46.30\textsubscript{$\pm$12.85} & 3.70\textsubscript{$\pm$5.76} \\
Attribute Perception & 83.78\textsubscript{$\pm$4.84} & 78.38\textsubscript{$\pm$5.39} & 18.92\textsubscript{$\pm$5.13} & 3.15\textsubscript{$\pm$2.41} & 33.33\textsubscript{$\pm$6.16} & 5.86\textsubscript{$\pm$3.15} \\
Action Recognition & 71.25\textsubscript{$\pm$4.99} & 67.09\textsubscript{$\pm$5.18} & 16.29\textsubscript{$\pm$4.08} & 2.56\textsubscript{$\pm$1.83} & 30.03\textsubscript{$\pm$5.05} & 5.43\textsubscript{$\pm$2.55} \\
Object Recognition & 76.27\textsubscript{$\pm$4.42} & 71.75\textsubscript{$\pm$4.67} & 17.80\textsubscript{$\pm$3.98} & 2.26\textsubscript{$\pm$1.63} & 34.46\textsubscript{$\pm$4.93} & 10.45\textsubscript{$\pm$3.20} \\
OCR Problems & 77.70\textsubscript{$\pm$6.86} & 72.66\textsubscript{$\pm$7.34} & 12.23\textsubscript{$\pm$5.46} & 0.72\textsubscript{$\pm$1.92} & 34.53\textsubscript{$\pm$7.81} & 6.47\textsubscript{$\pm$4.21} \\
Counting Problem & 51.49\textsubscript{$\pm$5.94} & 48.88\textsubscript{$\pm$5.95} & 11.94\textsubscript{$\pm$3.89} & 4.85\textsubscript{$\pm$2.63} & 19.40\textsubscript{$\pm$4.72} & 8.58\textsubscript{$\pm$3.38} \\
Temporal Reasoning & 67.80\textsubscript{$\pm$6.82} & 55.93\textsubscript{$\pm$7.23} & 5.08\textsubscript{$\pm$3.34} & 5.08\textsubscript{$\pm$3.34} & 10.73\textsubscript{$\pm$4.59} & 5.08\textsubscript{$\pm$3.34} \\
Spatial Reasoning & 80.36\textsubscript{$\pm$10.25} & 75.00\textsubscript{$\pm$11.09} & 25.00\textsubscript{$\pm$11.08} & 3.57\textsubscript{$\pm$5.57} & 60.71\textsubscript{$\pm$12.39} & 10.71\textsubscript{$\pm$8.23} \\
Action Reasoning & 64.21\textsubscript{$\pm$5.53} & 60.35\textsubscript{$\pm$5.64} & 11.23\textsubscript{$\pm$3.67} & 4.91\textsubscript{$\pm$2.56} & 15.79\textsubscript{$\pm$4.23} & 7.37\textsubscript{$\pm$3.06} \\
Object Reasoning & 69.60\textsubscript{$\pm$4.22} & 65.20\textsubscript{$\pm$4.36} & 13.22\textsubscript{$\pm$3.12} & 3.08\textsubscript{$\pm$1.63} & 21.81\textsubscript{$\pm$3.79} & 6.61\textsubscript{$\pm$2.30} \\
Information Synopsis & 86.38\textsubscript{$\pm$3.74} & 81.73\textsubscript{$\pm$4.20} & 16.72\textsubscript{$\pm$4.06} & 3.72\textsubscript{$\pm$2.12} & 29.10\textsubscript{$\pm$4.93} & 8.36\textsubscript{$\pm$3.04} \\
\textbf{Overall} & \textbf{72.33}\textsubscript{$\pm$1.69} & \textbf{67.56}\textsubscript{$\pm$1.77} & \textbf{14.33}\textsubscript{$\pm$1.32} & \textbf{3.44}\textsubscript{$\pm$0.69} & \textbf{26.63}\textsubscript{$\pm$1.67} & \textbf{7.37}\textsubscript{$\pm$0.99} \\
\bottomrule
\end{tabular}
\end{table*}

\begin{table*}[h]
\caption{\textbf{Per-task-type 1-bit accuracy (\%) on Video-MME (no-subtitle) for LLaVA-OneVision-7B ($N{=}2{,}700$).} Each entry is the point estimate followed by the $95\%$ CI half-width in subscript. The Overall row is bolded.}
\label{tab:subtype_videomme_lov}
\centering
\scriptsize
\setlength{\tabcolsep}{4pt}
\renewcommand{\arraystretch}{1.0}
\begin{tabular}{p{0.20\linewidth} *{6}{P{0.105\linewidth}}}
\toprule
Task type & bf16 & TACT-KV & TQ & KIVI & VidKV & AKVQ-VL \\
\midrule
Temporal Perception & 61.82\textsubscript{$\pm$12.44} & 54.55\textsubscript{$\pm$12.72} & 25.45\textsubscript{$\pm$11.24} & 9.09\textsubscript{$\pm$7.81} & 32.73\textsubscript{$\pm$12.04} & 18.18\textsubscript{$\pm$10.07} \\
Spatial Perception & 61.11\textsubscript{$\pm$12.58} & 51.85\textsubscript{$\pm$12.88} & 53.70\textsubscript{$\pm$12.85} & 11.11\textsubscript{$\pm$8.50} & 55.56\textsubscript{$\pm$12.81} & 29.63\textsubscript{$\pm$11.84} \\
Attribute Perception & 75.68\textsubscript{$\pm$5.61} & 69.82\textsubscript{$\pm$6.00} & 50.90\textsubscript{$\pm$6.52} & 9.46\textsubscript{$\pm$3.88} & 39.64\textsubscript{$\pm$6.39} & 22.07\textsubscript{$\pm$5.43} \\
Action Recognition & 54.95\textsubscript{$\pm$5.48} & 54.31\textsubscript{$\pm$5.48} & 37.38\textsubscript{$\pm$5.33} & 7.99\textsubscript{$\pm$3.02} & 37.06\textsubscript{$\pm$5.32} & 19.17\textsubscript{$\pm$4.35} \\
Object Recognition & 66.95\textsubscript{$\pm$4.88} & 58.76\textsubscript{$\pm$5.10} & 47.46\textsubscript{$\pm$5.17} & 10.17\textsubscript{$\pm$3.16} & 42.66\textsubscript{$\pm$5.12} & 20.62\textsubscript{$\pm$4.21} \\
OCR Problems & 60.43\textsubscript{$\pm$8.03} & 54.68\textsubscript{$\pm$8.16} & 37.41\textsubscript{$\pm$7.94} & 5.76\textsubscript{$\pm$4.00} & 42.45\textsubscript{$\pm$8.11} & 18.71\textsubscript{$\pm$6.45} \\
Counting Problem & 37.69\textsubscript{$\pm$5.76} & 35.07\textsubscript{$\pm$5.68} & 29.10\textsubscript{$\pm$5.41} & 4.48\textsubscript{$\pm$2.54} & 28.73\textsubscript{$\pm$5.39} & 11.57\textsubscript{$\pm$3.84} \\
Temporal Reasoning & 42.94\textsubscript{$\pm$7.21} & 42.37\textsubscript{$\pm$7.21} & 21.47\textsubscript{$\pm$6.02} & 7.34\textsubscript{$\pm$3.91} & 24.29\textsubscript{$\pm$6.27} & 14.12\textsubscript{$\pm$5.13} \\
Spatial Reasoning & 82.14\textsubscript{$\pm$9.92} & 78.57\textsubscript{$\pm$10.55} & 64.29\textsubscript{$\pm$12.18} & 5.36\textsubscript{$\pm$6.38} & 58.93\textsubscript{$\pm$12.47} & 21.43\textsubscript{$\pm$10.55} \\
Action Reasoning & 54.04\textsubscript{$\pm$5.75} & 49.47\textsubscript{$\pm$5.77} & 32.28\textsubscript{$\pm$5.39} & 8.42\textsubscript{$\pm$3.25} & 38.60\textsubscript{$\pm$5.61} & 21.05\textsubscript{$\pm$4.71} \\
Object Reasoning & 56.39\textsubscript{$\pm$4.55} & 54.63\textsubscript{$\pm$4.56} & 36.12\textsubscript{$\pm$4.40} & 5.95\textsubscript{$\pm$2.19} & 39.43\textsubscript{$\pm$4.48} & 17.40\textsubscript{$\pm$3.49} \\
Information Synopsis & 73.99\textsubscript{$\pm$4.77} & 72.14\textsubscript{$\pm$4.86} & 51.70\textsubscript{$\pm$5.41} & 8.05\textsubscript{$\pm$2.99} & 54.49\textsubscript{$\pm$5.40} & 21.98\textsubscript{$\pm$4.50} \\
\textbf{Overall} & \textbf{59.26}\textsubscript{$\pm$1.85} & \textbf{55.63}\textsubscript{$\pm$1.87} & \textbf{39.56}\textsubscript{$\pm$1.84} & \textbf{7.63}\textsubscript{$\pm$1.00} & \textbf{40.00}\textsubscript{$\pm$1.85} & \textbf{18.96}\textsubscript{$\pm$1.48} \\
\bottomrule
\end{tabular}
\end{table*}

\begin{table*}[h]
\caption{\textbf{Per-task-type 1-bit accuracy (\%) on Video-MME (no-subtitle) for VideoLLaMA3-7B ($N{=}2{,}700$).} Each entry is the point estimate followed by the $95\%$ CI half-width in subscript. The Overall row is bolded.}
\label{tab:subtype_videomme_v37b}
\centering
\scriptsize
\setlength{\tabcolsep}{4pt}
\renewcommand{\arraystretch}{1.0}
\begin{tabular}{p{0.20\linewidth} *{6}{P{0.105\linewidth}}}
\toprule
Task type & bf16 & TACT-KV & TQ & KIVI & VidKV & AKVQ-VL \\
\midrule
Temporal Perception & 65.45\textsubscript{$\pm$12.20} & 65.45\textsubscript{$\pm$12.20} & 38.18\textsubscript{$\pm$12.44} & 9.09\textsubscript{$\pm$7.81} & 40.00\textsubscript{$\pm$12.53} & 12.73\textsubscript{$\pm$8.86} \\
Spatial Perception & 64.81\textsubscript{$\pm$12.35} & 62.96\textsubscript{$\pm$12.47} & 51.85\textsubscript{$\pm$12.88} & 9.26\textsubscript{$\pm$7.95} & 48.15\textsubscript{$\pm$12.88} & 22.22\textsubscript{$\pm$10.87} \\
Attribute Perception & 76.58\textsubscript{$\pm$5.55} & 72.07\textsubscript{$\pm$5.87} & 44.59\textsubscript{$\pm$6.48} & 9.91\textsubscript{$\pm$3.96} & 40.99\textsubscript{$\pm$6.42} & 13.06\textsubscript{$\pm$4.44} \\
Action Recognition & 67.09\textsubscript{$\pm$5.18} & 64.54\textsubscript{$\pm$5.27} & 41.53\textsubscript{$\pm$5.43} & 5.75\textsubscript{$\pm$2.62} & 39.62\textsubscript{$\pm$5.39} & 13.42\textsubscript{$\pm$3.78} \\
Object Recognition & 71.19\textsubscript{$\pm$4.70} & 66.95\textsubscript{$\pm$4.88} & 40.68\textsubscript{$\pm$5.09} & 8.47\textsubscript{$\pm$2.92} & 39.27\textsubscript{$\pm$5.06} & 14.69\textsubscript{$\pm$3.69} \\
OCR Problems & 73.38\textsubscript{$\pm$7.27} & 65.47\textsubscript{$\pm$7.81} & 42.45\textsubscript{$\pm$8.11} & 5.76\textsubscript{$\pm$4.00} & 38.13\textsubscript{$\pm$7.97} & 13.67\textsubscript{$\pm$5.71} \\
Counting Problem & 44.78\textsubscript{$\pm$5.91} & 42.54\textsubscript{$\pm$5.88} & 29.48\textsubscript{$\pm$5.43} & 8.96\textsubscript{$\pm$3.45} & 34.33\textsubscript{$\pm$5.65} & 10.45\textsubscript{$\pm$3.68} \\
Temporal Reasoning & 50.28\textsubscript{$\pm$7.29} & 45.20\textsubscript{$\pm$7.25} & 24.86\textsubscript{$\pm$6.32} & 6.21\textsubscript{$\pm$3.63} & 22.03\textsubscript{$\pm$6.07} & 7.91\textsubscript{$\pm$4.04} \\
Spatial Reasoning & 82.14\textsubscript{$\pm$9.92} & 75.00\textsubscript{$\pm$11.09} & 58.93\textsubscript{$\pm$12.47} & 3.57\textsubscript{$\pm$5.57} & 62.50\textsubscript{$\pm$12.29} & 17.86\textsubscript{$\pm$9.92} \\
Action Reasoning & 58.95\textsubscript{$\pm$5.68} & 59.65\textsubscript{$\pm$5.66} & 38.95\textsubscript{$\pm$5.62} & 8.07\textsubscript{$\pm$3.19} & 37.89\textsubscript{$\pm$5.59} & 10.53\textsubscript{$\pm$3.58} \\
Object Reasoning & 60.35\textsubscript{$\pm$4.48} & 57.27\textsubscript{$\pm$4.53} & 39.87\textsubscript{$\pm$4.48} & 6.39\textsubscript{$\pm$2.27} & 40.09\textsubscript{$\pm$4.49} & 11.01\textsubscript{$\pm$2.89} \\
Information Synopsis & 79.88\textsubscript{$\pm$4.37} & 76.78\textsubscript{$\pm$4.59} & 53.56\textsubscript{$\pm$5.41} & 7.43\textsubscript{$\pm$2.89} & 56.97\textsubscript{$\pm$5.36} & 13.31\textsubscript{$\pm$3.71} \\
\textbf{Overall} & \textbf{65.19}\textsubscript{$\pm$1.80} & \textbf{62.00}\textsubscript{$\pm$1.83} & \textbf{40.81}\textsubscript{$\pm$1.85} & \textbf{7.44}\textsubscript{$\pm$0.99} & \textbf{40.56}\textsubscript{$\pm$1.85} & \textbf{12.44}\textsubscript{$\pm$1.24} \\
\bottomrule
\end{tabular}
\end{table*}

\section{Additional Baseline Comparisons}
\label{app:cr_baselines}
We compare VL-Cache~\cite{tu2024vl} and FlashCache~\cite{yang2025revisiting} at the measured visual-cache footprint of TACT-KV at a nominal 1-bit rate. Their retention budgets are set separately for each model and dataset. Table~\ref{tab:cr_eviction} reports macro prediction-flip rates relative to bf16. Both methods retain non-visual tokens at bf16 and compress the visual cache after prefill. VL-Cache and FlashCache achieve lower flip rates in five of the twelve model and dataset pairs. TACT-KV has the lowest flip rate in the other seven.

\begin{table}[t]
\caption{\textbf{Macro prediction-flip rate (\%) relative to bf$16$ on RSNA-ICH and CT-RATE at matched visual-cache footprints.} Retention budgets match the measured footprint of TACT-KV at a nominal 1-bit rate. Entries include Wald $95\%$ CI half-widths. \textbf{Bold} and \underline{underline} mark the lowest and second-lowest values in each column within a model.}
\label{tab:cr_eviction}
\centering\scriptsize
\setlength{\tabcolsep}{5pt}
\renewcommand{\arraystretch}{1.05}
\begin{tabular}{l *{2}{P{0.21\linewidth}}}
\toprule
Method & \textbf{RSNA-ICH} & \textbf{CT-RATE} \\
\midrule
\rowcolor{vlmgray}\multicolumn{3}{c}{\textbf{\textit{MedGemma 1.5 4B}}}\\
TACT-KV (ours) & \underline{3.78}\textsubscript{$\pm$0.28} & \underline{4.69}\textsubscript{$\pm$0.18} \\
VL-Cache & \textbf{2.77}\textsubscript{$\pm$0.24} & \textbf{4.21}\textsubscript{$\pm$0.17} \\
FlashCache & 8.27\textsubscript{$\pm$0.40} & 9.35\textsubscript{$\pm$0.24} \\
\rowcolor{vlmgray}\multicolumn{3}{c}{\textbf{\textit{MedGemma 4B}}}\\
TACT-KV (ours) & \underline{2.37}\textsubscript{$\pm$0.22} & \textbf{2.95}\textsubscript{$\pm$0.14} \\
VL-Cache & \textbf{2.17}\textsubscript{$\pm$0.21} & \underline{3.58}\textsubscript{$\pm$0.16} \\
FlashCache & 7.77\textsubscript{$\pm$0.39} & 8.58\textsubscript{$\pm$0.23} \\
\rowcolor{vlmgray}\multicolumn{3}{c}{\textbf{\textit{MedGemma 27B}}}\\
TACT-KV (ours) & 0.37\textsubscript{$\pm$0.09} & \textbf{2.63}\textsubscript{$\pm$0.13} \\
VL-Cache & \underline{0.33}\textsubscript{$\pm$0.08} & \underline{3.57}\textsubscript{$\pm$0.16} \\
FlashCache & \textbf{0.23}\textsubscript{$\pm$0.07} & 6.02\textsubscript{$\pm$0.20} \\
\rowcolor{vlmgray}\multicolumn{3}{c}{\textbf{\textit{Hulu-Med 7B}}}\\
TACT-KV (ours) & \textbf{6.08}\textsubscript{$\pm$0.35} & \textbf{10.44}\textsubscript{$\pm$0.26} \\
VL-Cache & \underline{13.49}\textsubscript{$\pm$0.50} & \underline{15.91}\textsubscript{$\pm$0.31} \\
FlashCache & 14.01\textsubscript{$\pm$0.51} & 19.41\textsubscript{$\pm$0.33} \\
\rowcolor{vlmgray}\multicolumn{3}{c}{\textbf{\textit{Hulu-Med 4B}}}\\
TACT-KV (ours) & \textbf{8.53}\textsubscript{$\pm$0.41} & \textbf{7.22}\textsubscript{$\pm$0.22} \\
VL-Cache & 10.30\textsubscript{$\pm$0.44} & \underline{11.60}\textsubscript{$\pm$0.27} \\
FlashCache & \underline{8.84}\textsubscript{$\pm$0.41} & 14.72\textsubscript{$\pm$0.30} \\
\rowcolor{vlmgray}\multicolumn{3}{c}{\textbf{\textit{Hulu-Med 32B}}}\\
TACT-KV (ours) & \underline{9.13}\textsubscript{$\pm$0.42} & \textbf{10.04}\textsubscript{$\pm$0.25} \\
VL-Cache & 11.53\textsubscript{$\pm$0.47} & \underline{11.27}\textsubscript{$\pm$0.27} \\
FlashCache & \textbf{8.85}\textsubscript{$\pm$0.41} & 13.17\textsubscript{$\pm$0.28} \\
\bottomrule
\end{tabular}
\end{table}

\subsection{Implementation Details}
All baselines use the same visual inputs and prompts as TACT-KV. Non-visual tokens and newly generated tokens remain at bf16. KVQuant uses calibrated nonuniform codebooks, pre-RoPE key quantization, and a 1\% outlier fraction stored at fp16. We construct codebooks at 1, 2, 3, and 4 bits. KVTC uses the public implementation, with PCA bases fitted per layer and attention head, adaptive bit allocation, affine scalar quantization, and DEFLATE entropy coding. Four sink tokens and the most recent 128 tokens remain at bf16. A nominal budget of $b$ bits corresponds to an average of $b$ bits per PCA coefficient before entropy coding.

Calibration uses general web text. We use 80 windows of 2,048 tokens, with 16 windows used to fit KVQuant codebooks and 512-token windows for its Fisher estimates. The calibration data are shared across bit budgets. VL-Cache allocates retention budgets using attention scores. FlashCache uses a token-axis DCT and outlier retention, with its frequency parameter set to $0.2$.

\section{Closure at Higher Bit Rates}
\label{app:cr_closure}
We extend the closure check in Table~\ref{tab:closure} to 2- and 4-bit average rates for all twelve VLM-dataset pairs. Table~\ref{tab:cr_closure} reports the median K-cache reconstruction-MSE ratio $\mathcal{D}_{\mathrm{uniform}}/\mathcal{D}_{\mathrm{water\text{-}fill}}$ across study-layer measurements, with the transform and quantizer held fixed. Eight of twelve pairs are within 30\% of the prediction at 2 bits, and eleven are within 30\% at 4 bits. The remaining 4-bit pair is within 50\%.

\begin{table}[t]
\caption{\textbf{Theorem~\ref{thm:water_fill} closure on twelve VLM-dataset pairs at 2- and 4-bit average rates.} For each pair we report the closed-form prediction $1/\gamma_K$ and the empirical ratios at each rate. \textbf{Bold} marks closure within $30\%$ of $1/\gamma_K$, \underline{underline} within $50\%$.}
\label{tab:cr_closure}
\centering\small
\setlength{\tabcolsep}{6pt}
\begin{tabular}{l *{3}{P{0.09\linewidth}} | *{3}{P{0.09\linewidth}}}
\toprule
& \multicolumn{3}{c|}{RSNA-ICH} & \multicolumn{3}{c}{CT-RATE} \\
\cmidrule(lr){2-4}\cmidrule(lr){5-7}
Model & $1/\gamma_K$ & 2-bit & 4-bit & $1/\gamma_K$ & 2-bit & 4-bit \\
\midrule
MedGemma 1.5 4B & 3.58 & \textbf{2.97} & \textbf{2.65} & 5.85 & \underline{8.76} & \textbf{7.02} \\
MedGemma 4B & 3.06 & \textbf{2.84} & \textbf{2.45} & 5.00 & \underline{7.05} & \textbf{5.80} \\
MedGemma 27B & 3.77 & \textbf{2.82} & \underline{2.61} & 5.56 & \textbf{6.08} & \textbf{5.42} \\
Hulu-Med 7B & 4.85 & \textbf{4.01} & \textbf{3.61} & 7.58 & \underline{10.35} & \textbf{8.45} \\
Hulu-Med 4B & 3.95 & \textbf{3.23} & \textbf{3.04} & 6.17 & \underline{9.01} & \textbf{7.99} \\
Hulu-Med 32B & 4.46 & \textbf{3.75} & \textbf{3.36} & 6.58 & \textbf{7.51} & \textbf{6.43} \\
\bottomrule\end{tabular}\end{table}

\section{Reconstruction Error of the Inner Scalar Quantizer}
\label{app:quant_mechanism}

This appendix gives a layer-level mechanistic account of the quantizer attribution result reported in Table~\ref{tab:abl_quantizer}. We compare three scalar quantizers, gain-shape SQ, symmetric uniform, and NormalFloat, on the actual post-DCT K-cache of MedGemma 1.5 4B at one transformer layer, isolating the inner scalar quantization step from the rest of the TACT-KV pipeline.

\smallskip
\noindent\textbf{Protocol.} We sample one RSNA-ICH study at the standard $S=28$ slice budget, run prefill on MedGemma 1.5 4B, and read out the K-cache at layer $5$ as a mid-stack representative. After the 3-axis DCT, the cache produces $S \cdot H \cdot W = 28 \cdot 16 \cdot 16 = 7{,}168$ frequency bins per (batch, head). The model has four KV heads at this layer, so we obtain $4 \times 7{,}168 = 28{,}672$ per-bin coefficient vectors of head dimension $d=256$. For each bit budget $b \in \{1, 2, 3, 4\}$ we quantize each vector with gain-shape SQ, symmetric uniform with per-vector absmax scaling, and NormalFloat~\cite{dettmers2023qlora}, then compute the dequantization error against the original coefficient vector. The DCT and the per-layer bit budget are held fixed across the three quantizers, so the only varying factor is the inner scalar quantizer.

\smallskip
\noindent\textbf{gain-shape SQ's reconstruction MSE is $168\times$ lower than the two naive baselines at $1$-bit.} Table~\ref{tab:quant_mse} reports per-quantizer reconstruction MSE at each bit budget. Relative to the per-tensor energy $\|F\|^2 / N = 1{,}722.97$, gain-shape SQ's $1$-bit reconstruction sits at $0.15\%$ error while Uniform sits at $25.6\%$. Uniform stays $21\times$ worse than gain-shape SQ at $4$-bit, where the model tolerates the residual magnitude and the downstream flip rate of Table~\ref{tab:abl_quantizer} converges, but at $1$-bit the same factor drives the catastrophic flip-rate degradation reported there.

\begin{table}[h]
\caption{\textbf{Per-quantizer reconstruction MSE on real post-DCT K-cache coefficients of MedGemma 1.5 4B.} Mean squared error between the original and dequantized coefficient vectors, computed over $28{,}672$ per-bin vectors at one transformer layer. Lower is better. The first row is the configuration used in TACT-KV.}
\label{tab:quant_mse}
\centering
\scriptsize
\setlength{\tabcolsep}{6pt}
\renewcommand{\arraystretch}{1.05}
\begin{tabular}{l *{4}{P{0.13\linewidth}}}
\toprule
Quantizer & $b{=}1$ & $b{=}2$ & $b{=}3$ & $b{=}4$ \\
\midrule
gain-shape SQ (TACT-KV)            & $\mathbf{2.617}$  & $\mathbf{0.768}$ & $\mathbf{0.218}$ & $\mathbf{0.0621}$ \\
Uniform sym              & $440.708$         & $43.189$         & $6.884$          & $1.288$ \\
NormalFloat                     & $440.708$         & $29.126$         & $3.373$          & $0.196$ \\
\bottomrule
\end{tabular}
\end{table}

\smallskip
\noindent\textbf{The MSE gap comes from the bulk of post-DCT coordinates that sit far below the per-vector absmax.} Table~\ref{tab:quant_mae_by_bin} partitions the $1$-bit reconstruction error by $|x|/\text{absmax}$. $60\%$ of all coordinates lie below $0.1\times$ the per-vector absmax, and symmetric uniform reconstructs every one of them at $\pm \text{absmax}$, yielding mean absolute error $6.114$. gain-shape SQ first extracts the per-vector $\ell_2$ norm $g$ and quantizes the rotated unit-vector residual, so the bulk is recovered at MAE $0.507$, a $12\times$ improvement on the dominant fraction of the data. The remaining $0.6\%$ of coordinates near the per-vector boundary trade in the opposite direction, where Uniform's $\pm \text{absmax}$ output happens to be close to truth and gain-shape SQ's residual is comparatively wide. Aggregated over the $7.34$ million coordinates, this trade produces the $168\times$ MSE gap of Table~\ref{tab:quant_mse}.

\begin{table}[h]
\caption{\textbf{$1$-bit reconstruction MAE on MedGemma 1.5 4B post-DCT K-cache, partitioned by per-coordinate magnitude.} Coordinates are binned by $|x| / \text{absmax}$ over the $28{,}672$ per-bin vectors. The first column reports the bin range, the second the coordinate count and its share of the total. The last column reports the multiplicative advantage of gain-shape SQ over Uniform.}
\label{tab:quant_mae_by_bin}
\centering
\scriptsize
\setlength{\tabcolsep}{6pt}
\renewcommand{\arraystretch}{1.05}
\begin{tabular}{l P{0.20\linewidth} P{0.13\linewidth} P{0.15\linewidth} P{0.16\linewidth}}
\toprule
$|x| / \text{absmax}$ range & \# coordinates (\%) & Uniform MAE & gain-shape SQ MAE & SQ advantage \\
\midrule
$[0.0, 0.1]$ (the bulk) & $4{,}396{,}691$ ($60.0\%$) & $6.114$ & $0.507$ & $12.1\times$ \\
$[0.1, 0.3]$            & $2{,}047{,}900$ ($28.0\%$) & $3.397$ & $0.405$ & $8.4\times$  \\
$[0.3, 0.5]$            & $\phantom{0,}569{,}408$ ($\phantom{0}7.8\%$)  & $2.445$ & $0.466$ & $5.2\times$  \\
$[0.5, 0.7]$            & $\phantom{0,}203{,}768$ ($\phantom{0}2.8\%$)  & $1.653$ & $0.567$ & $2.9\times$  \\
$[0.7, 0.9]$            & $\phantom{00,}76{,}210$ ($\phantom{0}1.0\%$)  & $0.871$ & $0.691$ & $1.3\times$  \\
$[0.9, 1.0]$            & $\phantom{00,}46{,}055$ ($\phantom{0}0.6\%$)  & $\mathbf{0.081}$ & $0.991$ & SQ $12\times$ worse \\
\bottomrule
\end{tabular}
\end{table}

\section{Tile-Fused Inverse-DCT Attention Kernel}
\label{app:kernel}

This appendix describes the fused inverse path referenced in Sections~\ref{sec:method:codebook} and~\ref{sec:exp:deployment}. The compressed K and V caches stored by TACT-KV live in the frequency domain as a tensor of $b_i$-bit packed indices and per-bin $\ell_2$-norm scalars. Decoding requires (i) dequantizing each frequency bin into the dequantized frequency tensor $\hat{F}$, (ii) applying the inverse three-axis DCT to recover the spatial-grid representation $\hat{X}$, and (iii) running the attention dot product against the current query. A naive serial implementation materializes the full reconstructed key tensor $\hat{X}$ of shape $B \times n_h \times n_v \times d$ and the corresponding value tensor at every decode step before launching attention. This intermediate footprint scales linearly with the decoding batch dimension $B$ and dominates the deployment-time HBM budget at large $B$, eliminating the iso-HBM headroom recovered by compression.

\smallskip
\noindent\textbf{The Q-projection commutes with the inverse three-axis DCT, so the full reconstructed cache is never needed.} The inverse three-axis DCT is a linear map. Writing $\Phi \in \mathbb{R}^{n_v \times n_v}$ for its composed matrix form (separable as $\Phi = D_s^{\!\top} \otimes D_h^{\!\top} \otimes D_w^{\!\top}$) and viewing $\hat{F}, \hat{X}$ as $(n_v \times d)$ matrices for one (batch, head) slot, we have $\hat{X} = \Phi \, \hat{F}$. For a single-row decoder query $Q \in \mathbb{R}^{d}$, the attention score against the $n$-th reconstructed key satisfies
\begin{equation*}
Q \cdot \hat{X}[n] \;=\; Q \cdot \big(\Phi \, \hat{F}\big)[n] \;=\; \sum_{n'=1}^{n_v} \Phi[n, n'] \, \big(Q \cdot \hat{F}[n']\big),
\end{equation*}
so the score equals a linear combination of the scalar projections of $Q$ against the \emph{frequency-domain} bins of $\hat{F}$, weighted by the $n$-th row of $\Phi$. The same identity applies on the value side with the attention weights in place of $Q$. The Q-projection and the weights-projection are therefore applied directly to the dequantized but un-transformed $\hat{F}$, producing a scalar per frequency bin, and the inverse three-axis DCT is applied last to those scalars rather than to the full $(n_v, d)$ tensor. The full key and value tensors are never instantiated.

\smallskip
\noindent\textbf{The kernel applies dequantization, the attention dot product, and the inverse DCT inline within each FlashAttention tile.} The identity above is realized as a Triton kernel in two stages. The first stage walks the frequency-domain bins in register-resident tiles. For each tile the kernel loads the packed index buffer at the bit width $b_i$, unpacks the indices, gathers the corresponding gain-shape SQ codebook entries, multiplies by the per-bin $\ell_2$-norm scalars, and accumulates the dot product against the query in fp32. The second stage performs the same per-tile dequantization on the value cache and accumulates the attention-weighted sum. The inverse three-axis DCT closes the kernel by a small matmul on the per-frequency scalars, which inherits the separability of the DCT and reduces to three sequential one-dimensional transforms along the slice, height, and width axes.

\smallskip
\noindent\textbf{Deployment-time HBM under the fused path is proportional to the compressed cache rather than to the reconstructed full cache.} The fused path replaces the $\mathcal{O}(B \cdot n_h \cdot n_v \cdot d)$ intermediate of the naive path with a per-tile scratch whose size is bounded by the FlashAttention block configuration and is independent of $B$. The HBM held during decode is therefore the compressed cache (packed indices plus per-bin $\ell_2$-norm scalars) plus a small per-tile transient, which is what enables the iso-HBM decoding-batch multipliers reported in Table~\ref{tab:deploy} for the Hulu-Med family.

\smallskip
\noindent\textbf{The fused path is implemented for Qwen2 and Qwen3 attention and transfers to other attention classes through the same per-tile principle.} The kernel is realized as a per-attention-class wrapper that replaces the model's attention forward on decode steps when the cache is a TACT-KV cache and the layer is full-attention. The Qwen2 and Qwen3 implementations cover the Hulu-Med family of vision-language models on which we report decoding-batch profiling in Section~\ref{sec:exp:deployment}. The same fusion identity transfers to other attention classes such as Gemma-3, Llama, and Qwen2.5-VL through their corresponding attention forward surfaces.

\section{Detailed Related Work}
\label{app:related}

This appendix provides extended related work.

\subsection*{LLM-Targeted KV Cache Quantization}

KIVI~\cite{liu2024kivi} quantizes the K cache per channel along token groups and the V cache per token along head-dimension groups under asymmetric uniform quantization at $2$-bit, on the observation that K exhibits channel-wise outliers while V exhibits token-wise magnitude variation. KVQuant~\cite{hooper2024kvquant} combines per-channel quantization on K with pre-RoPE rotation that removes the rotary embedding's effect on the quantization range, a sensitivity-weighted non-uniform codebook derived from a calibration set, and per-vector dense-and-sparse splitting that isolates outliers separately for each vector. TurboQuant~\cite{zandieh2025turboquant} rotates each per-token coordinate vector by a random orthonormal matrix that pushes the post-rotation distribution toward an isotropic Gaussian, then applies a Lloyd--Max scalar codebook tuned for that target distribution. Our gain-shape SQ codebook in Section~\ref{sec:method:codebook} inherits the rotation-then-Lloyd--Max pattern of TurboQuant Algorithm~1 and adds gain-shape extraction.

\subsection*{VLM-Targeted KV Cache Compression}

AKVQ-VL~\cite{su2025akvq} applies a Walsh--Hadamard transform along the head-dimension of K to suppress channel outliers, then per-channel asymmetric uniform quantization on the rotated K and per-token asymmetric uniform quantization on V, with an adaptive bit budget allocated by Text-Salient Attention (a layer-level selection of attention heads that concentrate on text tokens) and Pivot-Token-Salient Attention (a token-level rule that retains a small set of pivot tokens at higher precision). The published configuration is mixed-precision around an average of $2$-bit on the visual cache. VL-Cache~\cite{tu2024vl} is eviction-based rather than quantization-based, allocating a layer-adaptive sparsity budget across decoder layers and a modality-aware token scoring policy that distinguishes visual from text tokens during prefill and decoding, retaining about $10\%$ of the entries at full precision and discarding the rest.

\subsection*{Frequency-Domain Compression}

FreqKV~\cite{kai2025freqkv} targets context-window extension on text-only LLMs and applies a discrete cosine transform along the token-sequence axis. When the cache reaches a fixed maximum size, the method iteratively compresses older chunks in the frequency domain by retaining low-frequency components and discarding high-frequency components, on the observation that context information concentrates in the low-frequency band. Adapting the model to this representation requires a brief fine-tuning pass on a long-context corpus, in contrast to the calibration-free TACT-KV pipeline of Section~\ref{sec:method}. FlashCache~\cite{yang2025revisiting} extracts a low-frequency baseline of the multimodal KV cache by low-pass filtering and identifies outlier KV pairs as those that deviate substantially from this baseline. The compression retains the low-frequency baseline together with the outlier pairs and discards the rest. KVTC~\cite{staniszewski2025kv} is a transform-coding pipeline that applies a calibration-fitted PCA along the head dimension, an adaptive bit allocation across the principal components, and an entropy coder on the resulting indices. KVTC's basis is one-dimensional and is fitted per model from a calibration set, whereas TACT-KV uses a fixed three-axis DCT over the volumetric input geometry without calibration. Fourier-VLM~\cite{wang2025fourier} applies a $2$-D discrete cosine transform to the vision encoder's output features and discards high-frequency components by a low-pass filter before the compressed visual tokens enter the language-model decoder. The transform is therefore applied on the input side of the LLM and does not modify the per-layer KV cache, which is the target of TACT-KV.

\subsection*{Video-Domain KV Cache Compression}

Sali-Cache~\cite{sai2026dual} reuses cache entries across temporally adjacent video frames through optical-flow-based detection of inter-frame redundancy and applies a saliency-guided spatial filter that retains visually significant regions and prunes the rest. LOOK-M~\cite{wan2024look} is a plug-and-play eviction and merging policy for multimodal long-context inference. It observes that the model's attention during prompt prefill is biased toward text tokens, ranks visual cache entries by cumulative attention score with a text-side bonus, evicts the lowest-ranked entries, and merges the discarded entries into the surviving ones through averaged, pivotal, or similarity-weighted variants. The method is precision-preserving on the surviving entries and requires no fine-tuning. VidKV~\cite{tao2025plug} applies a $1$-D fast Fourier transform per token along the head-dimension of the K cache, then $1$-bit sign quantization on the real and imaginary parts of the resulting spectrum, with channels classified as anomalous held at $2$-bit asymmetric uniform instead of going through the FFT branch. The V cache uses $1.58$-bit ternary quantization with per-group absmax scaling along the head-dimension.

\section{Limitations }
\label{sec:limitations}

\smallskip\noindent\textbf{Limitations.}
The volumetric prior we exploit is validated on chest CT (CT-RATE) and head CT (RSNA-ICH); other 3D medical modalities such as MRI, PET, and 3D ultrasound carry the same anatomical-axis structure in principle but are not measured here. The closed form of Theorem~\ref{thm:water_fill} holds in the high-rate regime in which every frequency bin receives non-trivial bit allocation, and on the smaller Hulu-Med variants the lowest-rate operating point pushes the empirical ratio outside the band predicted by the theorem (Section~\ref{sec:exp:closure}); deriving a low-rate counterpart bound is left to future work. Our pipeline compresses the visual portion of the cache only and leaves text tokens at bf16, so the savings reported here are tied to inputs in which visual tokens dominate; integrating TACT-KV with attention-aware eviction on the text portion is orthogonal and a natural next step. At decode, the gain--shape codebook and the tile-fused inverse path add per-step compute relative to a uniform quantizer with no transform, so the iso-HBM batch capacity reported in Section~\ref{sec:exp:deployment} is recovered at the cost of additional FLOPs per decoded token; the trade-off favors HBM-bound deployment and is less attractive when decoding is FLOP-bound. Deployment profiling is reported on a single GPU class, and cross-hardware behavior at smaller HBM budgets is not measured.


\end{document}